\documentclass[hyperfootnotes=false]{article}

\usepackage{hyperref}
\usepackage{graphicx}
\usepackage[font=scriptsize]{subfig}
\usepackage{tabu}
\usepackage{multirow}
\usepackage{xcolor}
\usepackage{rotating}
\usepackage{MnSymbol}

\newcommand\blfootnote[1]{%
	\begingroup
	\renewcommand\thefootnote{}\footnote{#1}%
	\addtocounter{footnote}{-1}%
	\endgroup
}

\begin{document}
	
\title{STag: A Stable Fiducial Marker System}
	
\author{Burak Benligiray \and Cihan Topal \and Cuneyt Akinlar}

\date{}
	
\maketitle

\begin{abstract}
Fiducial markers provide better-defined features than the ones naturally available in the scene.
For this reason, they are widely utilized in computer vision applications where reliable pose estimation is required.
Factors such as imaging noise and subtle changes in illumination induce jitter on the estimated pose.
Jitter impairs robustness in vision and robotics applications, and deteriorates the sense of presence and immersion in AR/VR applications.
In this paper, we propose STag, a fiducial marker system that provides stable pose estimation.
STag is designed to be robust against jitter factors, thus sustains pose stability better than the existing solutions.
This is achieved by utilizing geometric features that can be localized more repeatably.
The outer square border of the marker is used for detection and homography estimation.
This is followed by a novel homography refinement step using the inner circular border.
After refinement, the pose can be estimated stably and robustly across viewing conditions.
These features are demonstrated with a comprehensive set of experiments, including comparisons with the state of the art fiducial marker systems.
\blfootnote{Accepted to be published in Image and Vision Computing.}
\blfootnote{Source code, supplementary video: \href{https://github.com/bbenligiray/stag}{https://github.com/bbenligiray/stag}.}
\end{abstract}

\section{Introduction}
\label{sec:introduction}

Fiducial markers are artificial patterns that combine fast and accurate pose estimation with easy and inexpensive deployment~\cite{Fiala:2010}.
These properties make them particularly useful for prototyping.
They were first proposed to be used for augmented reality~\cite{Rekimoto:1998,Kato:2000,Zhang:2002}, and are still relevant in the recent virtual reality boom~\cite{make:2016}.
In addition, they see increasing use in systems that require to interact precisely and responsively with the real world, including robotics~\cite{Olson:2011,amazon:2016,boston:2016}.

Stability is a key metric of pose estimation, and is especially critical for certain applications.
For example, when using the estimated pose in a mixed reality application, instability causes jittery graphics, which is more immersion-breaking than a constant misalignment.
Pose estimation instability is also highly undesirable in control applications, as it causes oscillatory behavior.
Therefore, fiducial marker systems should aim to improve stability along with the usual performance metrics, such as detection robustness, speed and library size.

The marker pose is estimated using the geometric feature localizations.
For better pose stability, these localizations have to be repeatable.
Therefore, using geometric features that can be localized better will improve pose stability.
In addition, a marker design must provide adequate projective constraints to estimate the pose.
To satisfy both of these criteria, we propose a combination of geometric features to be used as the marker design. 

\begin{figure}
	\centering
	\includegraphics[width=0.8\columnwidth]{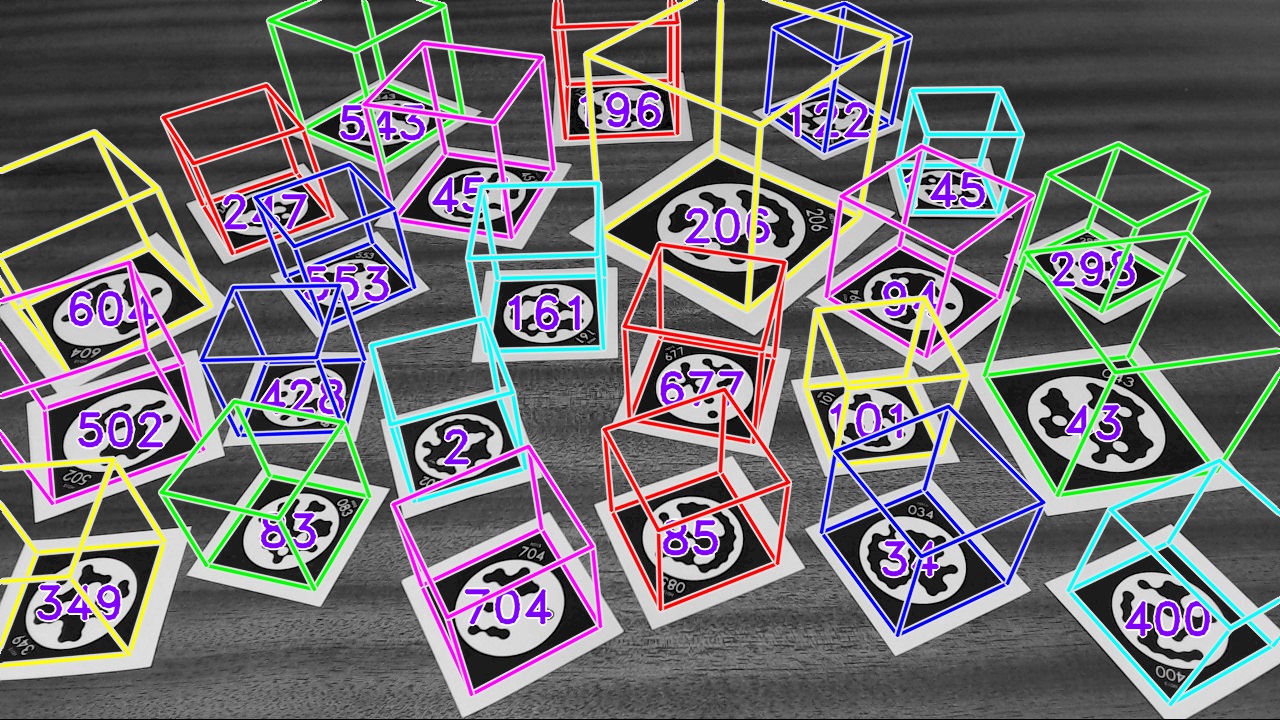}
	\caption{The poses estimated with the proposed marker are used to render 3D boxes.}
	\label{fig:teaser}
\end{figure}

In this paper, we propose STag, a fiducial marker system that provides stable localization without using any temporal filtering.
This is achieved by the hybrid marker design seen in Figure~\ref{fig:teaser}.
The outer square border is used for detection and homography estimation, and the inner circular border is used to refine the estimated homography.
Since both of these features are simple, they do not degrade when seen from long distances and acute viewing angles, resulting in robust detection.
The homography refinement step that provides stability is unique in that it uses a single conic correspondence to optimize the estimated homography.
To support the use of a wide range of numbers of markers, we generated various marker libraries by adapting the lexicographic generation algorithm.

%-------------------------------------------------------------------------
\section{Related Work}

\begin{figure}
	\def\mywidth{0.18\columnwidth}
	\centering
	\subfloat[ARToolkit~\cite{Kato:1999}]
	{
		\label{fig:litmarkers-artoolkit}
		\includegraphics[width=\mywidth]{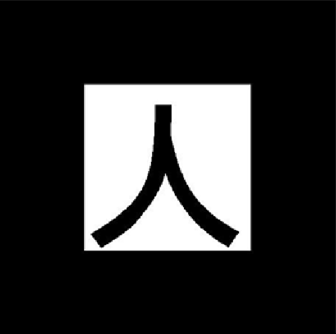}
	}
	\subfloat[ARTag~\cite{Fiala:2005a}]
	{
		\label{fig:litmarkers-artag}
		\includegraphics[width=\mywidth]{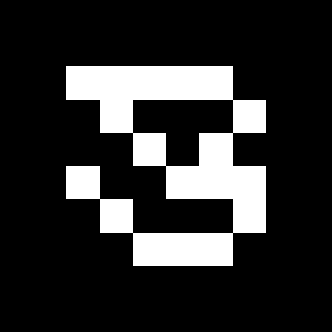}
	}
	\subfloat[AprilTag~\cite{Olson:2011}]
	{
		\label{fig:litmarkers-apriltag}
		\includegraphics[width=\mywidth]{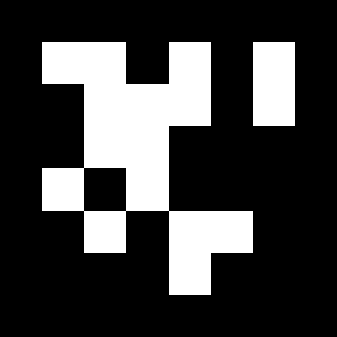}
	}
	\subfloat[ArUco~\cite{Garrido:2014}]
	{
		\label{fig:litmarkers-aruco}
		\includegraphics[width=\mywidth]{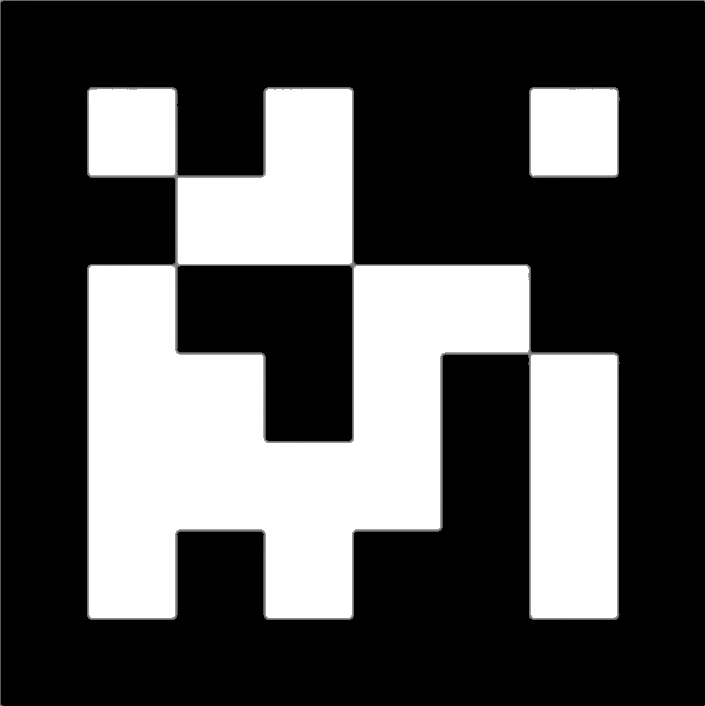}
	}
	
	\subfloat[TRIP~\cite{Lopez:2002}]
	{
		\label{fig:litmarkers-trip}
		\includegraphics[width=\mywidth]{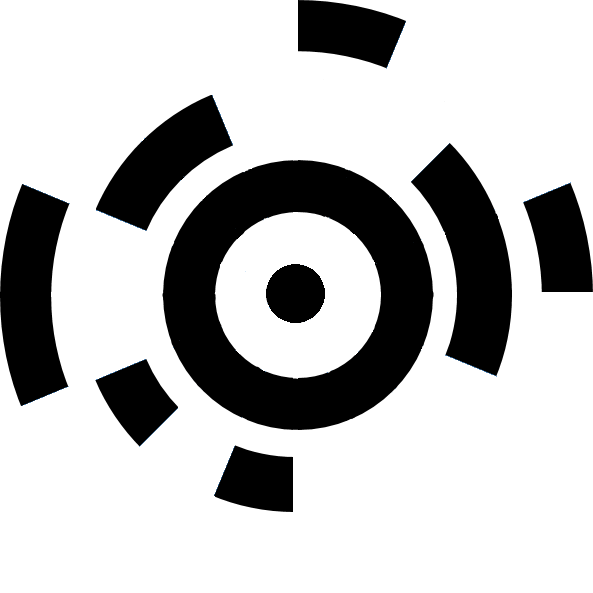}
	}
	\subfloat[RUNE-Tag~\cite{Bergamasco:2016}]
	{
		\label{fig:litmarkers-runetag}
		\includegraphics[width=\mywidth]{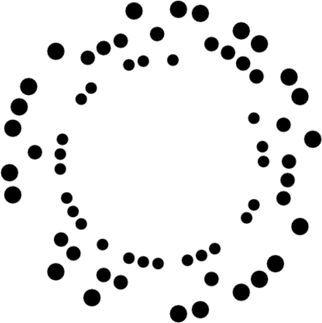}
	}
	\subfloat[ChromaTag~\cite{DeGol:2017}]
	{
		\label{fig:litmarkers-chroma}
		\includegraphics[width=\mywidth]{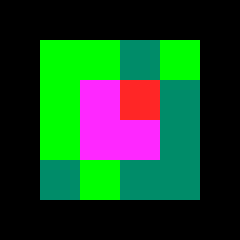}
	}
	\subfloat[CCTag~\cite{Calvet:2016}]
	{
		\label{fig:litmarkers-cctag}
		\includegraphics[width=\mywidth]{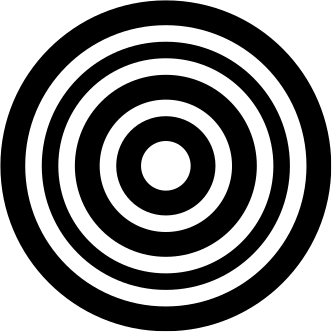}
	}
	\medskip
	\caption{Marker designs of some of the notable fiducial marker systems in the literature.}
	\label{fig:litmarkers}
\end{figure}

ARToolkit~\cite{Kato:1999} markers are encoded with patterns chosen by the user, each of which is associated with an ID (see Figure~\ref{fig:litmarkers-artoolkit}).
These patterns are densely sampled and decoded using a nearest neighbor method.
This approach has error detection and correction functionality, but its performance depends on the specific set of patterns in the library.
Matrix-like coding option is added to ARToolkit with an extension~\cite{Wagner:2007}.

ARTag~\cite{Fiala:2005a} is the first marker system with forward error correction capability based on digital coding methods (see Figure~\ref{fig:litmarkers-artag}).
This influenced the following studies to place a heavier emphasis on coding.
Additionally, ARTag is the first marker system to detect markers from edges.

AprilTag~\cite{Olson:2011} shares the square barcode type design (see Figure~\ref{fig:litmarkers-apriltag}), but the user chooses the size of the encoding grid.
Other recent marker systems also provide marker variants with different bandwidths~\cite{Bergamasco:2011,Garrido:2014}.
AprilTag is the first marker system to use lexicodes, greedily generated codes with error detection and correction capabilities.

ArUco~\cite{Garrido:2014} has the unique feature of calculating an occlusion mask with tiled markers, which can prove useful in mixed reality applications (see Figure~\ref{fig:litmarkers-aruco}).
The original ArUco marker library is created using a stochastic lexicode generation algorithm.
Better libraries are added using a mixed integer linear programming approach~\cite{Garrido:2016}.

TRIP~\cite{Lopez:2002} uses a circular design (see Figure~\ref{fig:litmarkers-trip}).
Markers have a ring in the middle for detection, and two outer rings for binary encoding.
Outer rings are divided into aligned slices.
The black slice pair indicates the starting point of the encoding.
Relying on singular features to resolve rotation ambiguity for circular designs is not unique to TRIP~\cite{Naimark:2002,Xu:2011}.
Having such regions is not preferable, as occluding them will prevent the marker from being detected.

RUNE-Tag~\cite{Bergamasco:2016,Bergamasco:2011} does not have a border in the traditional sense.
The coding dots are arranged in a circular shape, and act as candidate generating features (see Figure~\ref{fig:litmarkers-runetag}).
The pose is estimated by using the coding dots as point correspondences.
This approach results in a finite pose ambiguity similar to square markers, which can be resolved by decoding.
The finely detailed design provides many correspondence points, resulting in stable pose estimation.
However, these fine details are prone to degradation under suboptimal viewing conditions.

The literature tends to use monochromatic designs, as using colors will require certain photometric assumptions, or a calibration step~\cite{Fiala:2010}.
ChromaTag~\cite{DeGol:2017} modifies AprilTag to use colors for both candidate detection and decoding (see Figure~\ref{fig:litmarkers-chroma}).
The detection algorithm only runs on areas where both violet and green are present, which decreases the detection time significantly.

The systems we have mentioned until now are designed for pose estimation from a single marker, which is useful for applications such as augmented reality and robotics.
Square designs are dominant among these markers, as their corners provide the four correspondences required to estimate the homography.
Photogrammetry applications come with a slightly different requirement: Each marker needs to provide a single highly accurate correspondence.
In these applications, circular markers, and especially markers composed of concentric circles excel.

CCTag~\cite{Calvet:2016} is an improvement over the concentric circles commonly used in photogrammetry (see Figure~\ref{fig:litmarkers-cctag}).
Its main contribution is robustness against motion blur, an important issue in tracking applications.
Due to the reasons described above, CCTag does not estimate the pose of the marker, but only locates the projection of the marker center, similar to some earlier circular fiducial markers~\cite{Sattar:2007}.

%-------------------------------------------------------------------------
\section{Stable Marker Design}

To understand how stability can be improved, let us discuss what causes instability.
To estimate the pose of a fiducial marker, we localize its geometric features, and use them as correspondences.
Therefore, the uncertainty in the estimated pose is a result of the uncertainty in feature localizations.
We are going to investigate if we can improve stability further by utilizing geometric features that can be localized more certainly.

%-------------------------------------------------------------------------
\subsection{Geometric Features}
\label{sec:GeometricFeatures}

Various geometric features can be used to generate correspondences for pose estimation:
\begin{itemize}
	\item Lines are used in square markers~\cite{Fiala:2005a,Garrido:2014}.
	This is by far the most common solution in the literature, and does not have any obvious disadvantages.
	\item Points are used as correspondences in dot patterns~\cite{Uchiyama:2011,Bergamasco:2011}.
	These markers are composed of fine dots, which are difficult to detect robustly from acute viewing angles and long distances.
	\item Conics are used as correspondences in circular markers~\cite{Lopez:2002,Calvet:2016}.
	A single conic correspondence is not adequate to estimate the pose, which is why these markers either use additional geometric features~\cite{Lopez:2002}, or do not allow pose estimation with a single marker~\cite{Sattar:2007,Calvet:2016}.
\end{itemize}

We are going to argue that the traditional square border has a critical drawback compared to the circular border:
It cannot be localized as stably.
Therefore, circular borders should be utilized for utmost pose stability.
The problem of a single conic correspondence under-defining the pose still stands, to which we are going to propose a novel solution to in Section~\ref{sec:PoseRefinement}.

The square border of the fiducial marker appears as a quadrilateral (shortened as quad) on the image.
To localize the quad, we sample points from each vertex of the quad (e.g., by edge detection), and fit lines to individual point groups.
Either these lines, or their intersections can be used to estimate the marker pose.
Each line represents an exclusive quarter of the information that is used to estimate the pose, which means that the localization error of a line cannot be compensated for by the others.

The circular border of the fiducial marker appears as an ellipse on the image.
Differently from the quad, the ellipse cannot be represented as a combination of independent parts.
For this reason, a single fitting operation is applied to all samples.
This can be seen as the pose information being distributed to the entire shape, which allows it to be recovered more robustly.

\begin{figure}
	\centering
	\subfloat[Quad localization]{\label{fig:quadA}\includegraphics[width=0.4\columnwidth]{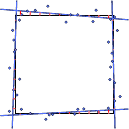}}
	\subfloat[Ellipse localization]{\label{fig:quadB}\includegraphics[width=0.4\columnwidth]{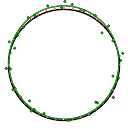}}
	\medskip
	\caption{A square and a circle are uniformly sampled at $40$ points, which are perturbed with a Gaussian noise with $0.04$ standard deviation along each axis.
		A quad is fitted to the samples from the square, and an ellipse is fitted to the samples from the circle.
		Localization errors are shown with red lines.}
	\label{fig:stoc}
\end{figure}

Let us consider a square and a circle, both centered on the origin with an area of $1$.
We uniformly sample $40$ points along these shapes, and perturb them with Gaussian noise.
The quad is localized by fitting a line to each $10$ samples with least squares.
The ellipse is localized by fitting an ellipse to all samples with a least squares method~\cite{Fitzgibbon:1999}.
To measure the mean localization error, the original shape is uniformly sampled at $40$ points, and the average distance of these samples to the fitted shape is calculated (see Figure~\ref{fig:stoc}).

\begin{figure}
	\centering
	\includegraphics[width=0.7\columnwidth]{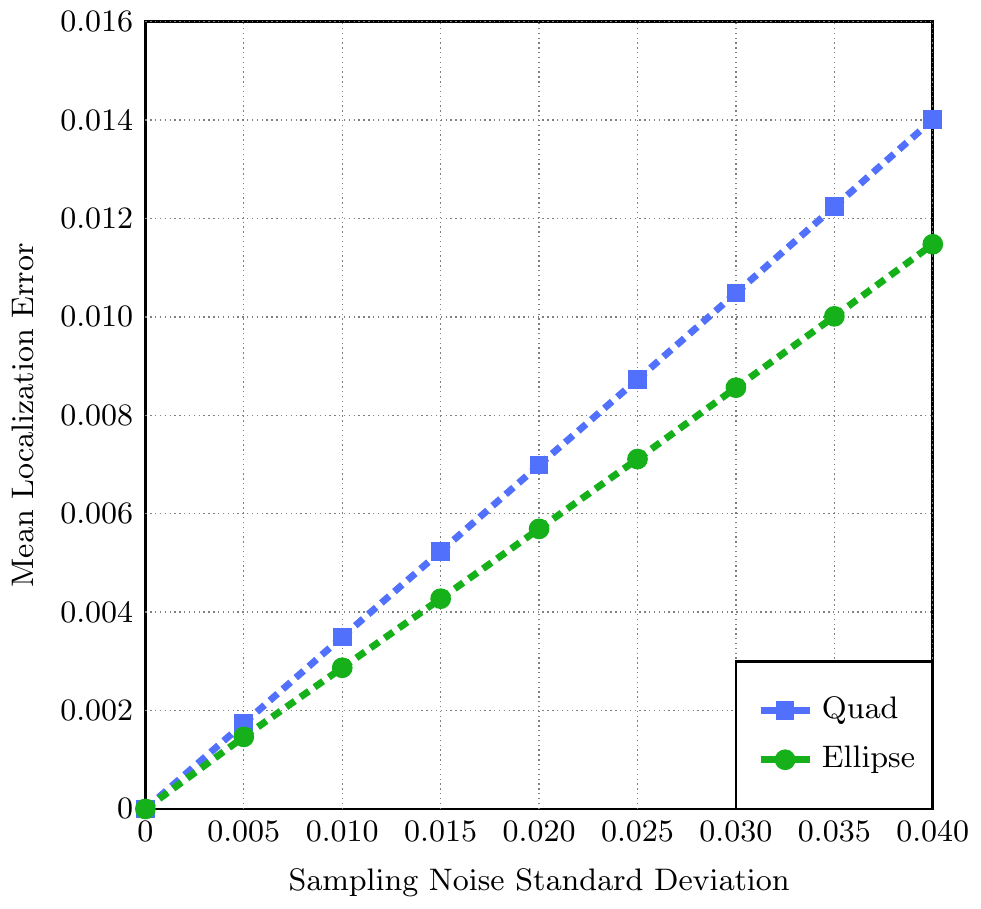}
	\caption{Mean localization errors with varying sampling noise standard deviation.
		Ellipse is localized better across all parameters}
	\label{fig:localization}
\end{figure}

The described experiment is run $10{,}000$ times with standard deviation values between $0$ and $0.04$.
See Figure~\ref{fig:localization} for the results.
As expected, the ellipse is localized better than the quad for all standard deviation values.
See Section~\ref{sec:poserefbenefit} for an experiment with real images.

%-------------------------------------------------------------------------
\subsection{Proposed Marker Design}

\begin{figure}
	\def\mywidth{0.25\columnwidth}
	\centering
	\subfloat[]
	{
		\label{fig:codetiling}
		\includegraphics[width=\mywidth]{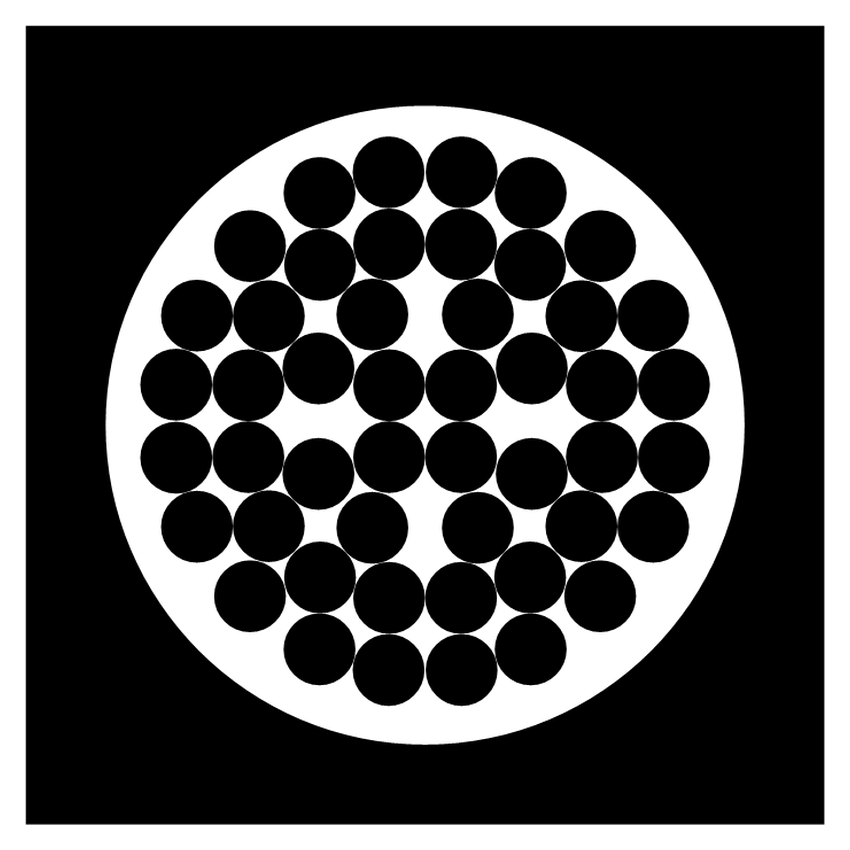}
	}
	\subfloat[]
	{
		\label{fig:plainencoding}
		\includegraphics[width=\mywidth]{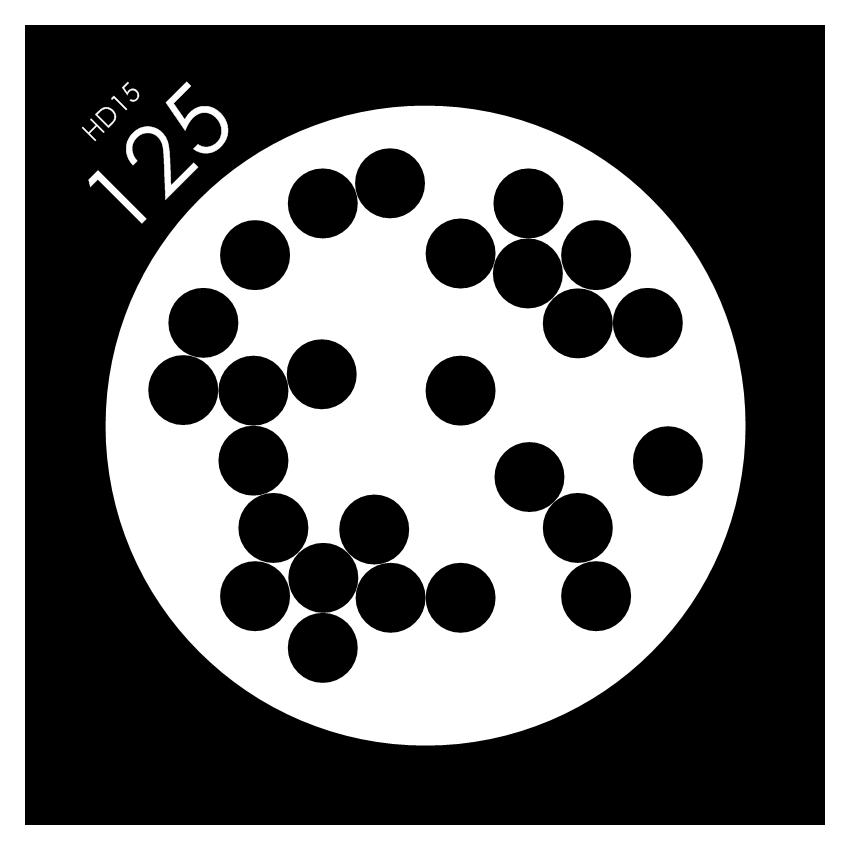}
	}
	\subfloat[]
	{
		\label{fig:morphencoding}
		\includegraphics[width=\mywidth]{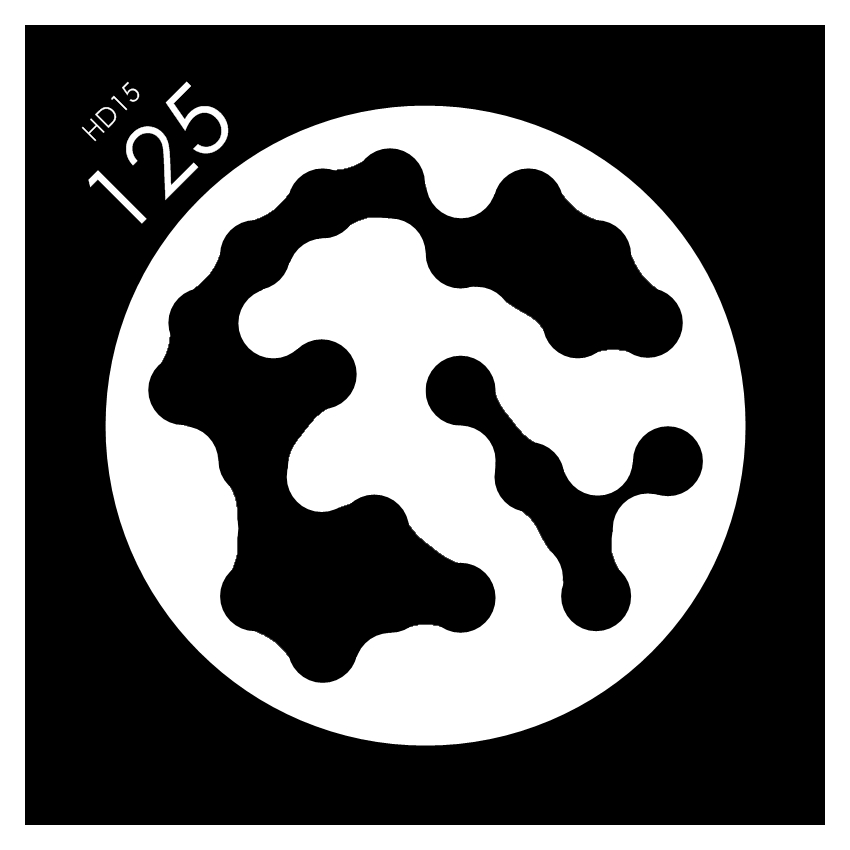}
	}
	\vspace{-0.5\baselineskip}
	\subfloat[]
	{
		\label{fig:markerexample}
		\includegraphics[width=\mywidth]{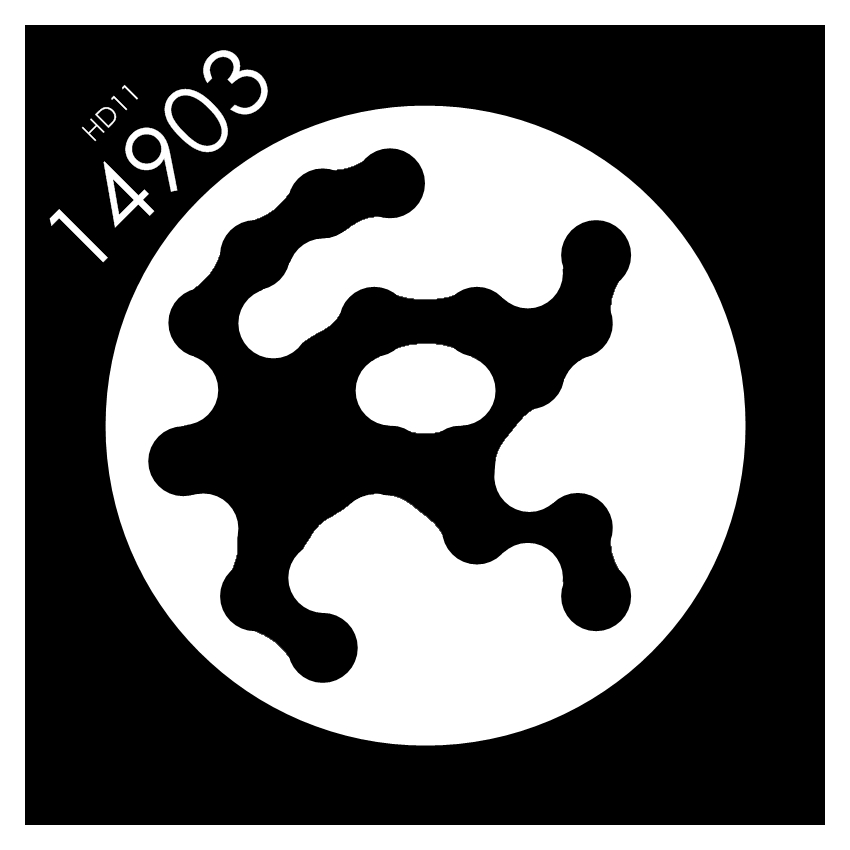}
	}
	\subfloat[]
	{
		\includegraphics[width=\mywidth]{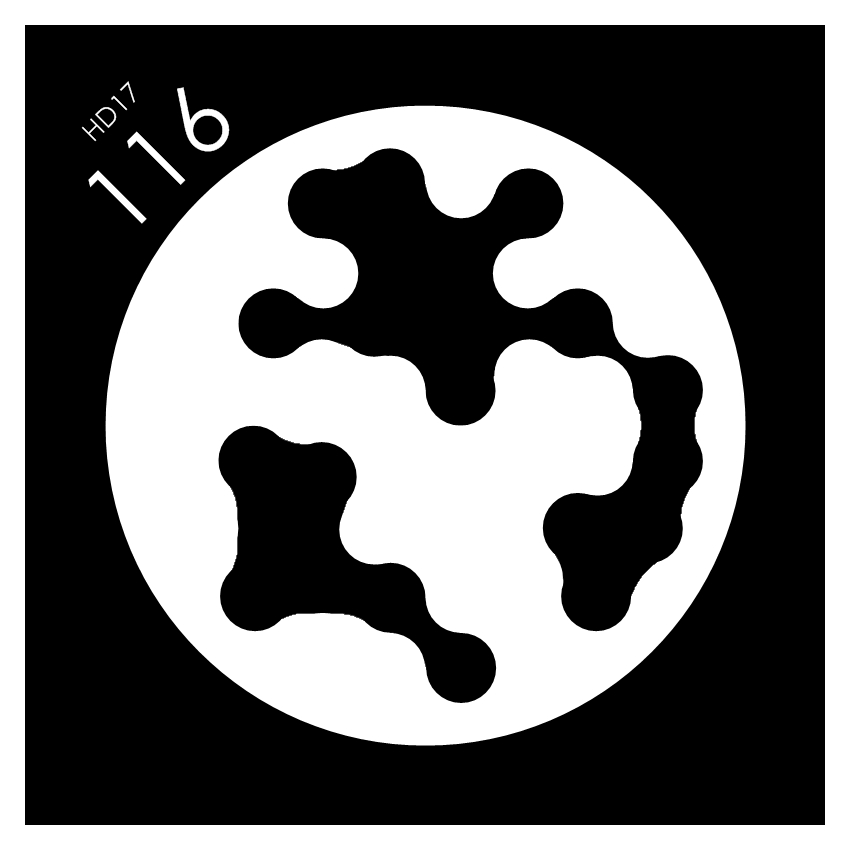}
	}
	\subfloat[]
	{
		\includegraphics[width=\mywidth]{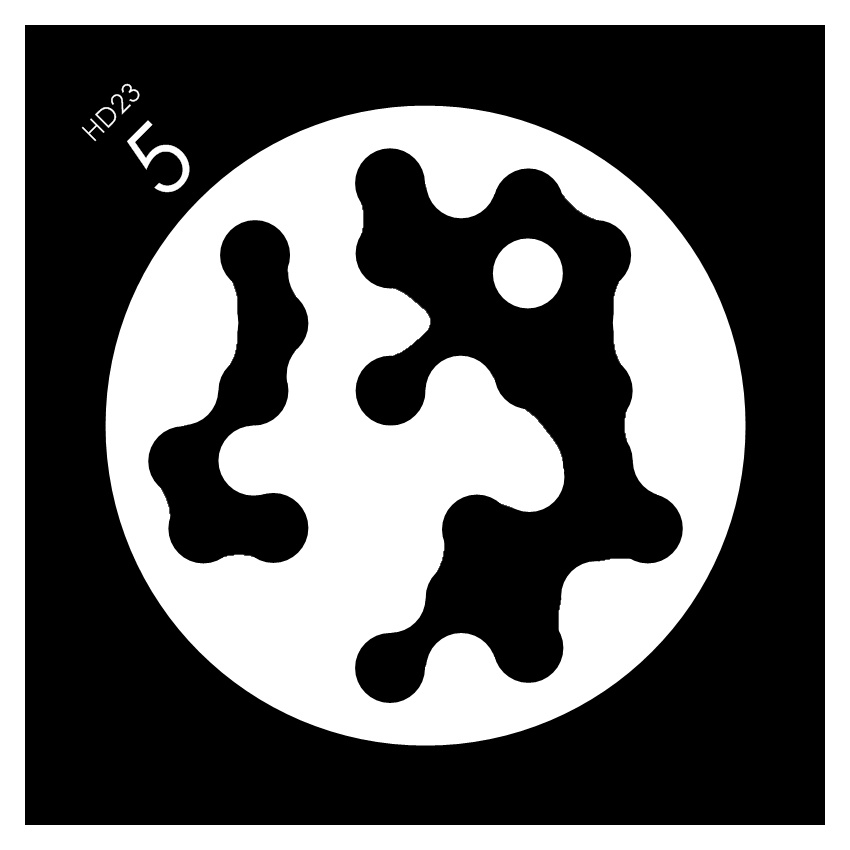}
	}
	
	\caption{(a) shows the tiling of bit representations in the marker.
		(b) and (c) are markers before and after morphological operations, respectively.
		(d)-(f) are example markers from libraries with different minimum Hamming distances.}
	\label{fig:design}
\end{figure}

In Section~\ref{sec:GeometricFeatures}, we argued that a circular border can be localized more stably.
By this motivation, the proposed stable marker design contains an inner circular border.
Since a single conic correspondence is not adequate to estimate the pose, additional geometric features need to be used.
An outer square border supplies the additional constraints, as seen in Figure~\ref{fig:design}.

The outer square border of the proposed marker is utilized to detect the marker and estimate a rough homography.
After detecting the marker and validating it with its encoding, the circular border is localized and used to refine the previously estimated homography.
The encoding area inside the inner circular border is filled with disk-shaped bit representations, similar to~\cite{Zhang:2002}.
48 disks are packed efficiently inside the circular area using a simulated annealing method~\cite{Kirkpatrick:1983} (see Figure~\ref{fig:codetiling}).

After encoding the marker as in Figure~\ref{fig:plainencoding}, the code pattern is morphologically dilated and eroded repeatedly, resulting in the encoding pattern in Figure~\ref{fig:morphencoding}.
Filling the gaps between neighboring bit representations allows the code to be read correctly in the case of slight localization errors.
It also reduces high frequency elements in the pattern, resulting in less edge detections.
Finally, it improves robustness against blooming and reflection effects, especially when the printing material is glossy.
See Figure~\hyperref[fig:markerexample]{5d-f} for additional example markers from libraries with different minimum Hamming distances.

%-------------------------------------------------------------------------
\subsection{Encoding}

We generated lexicographic marker libraries, as first done in~\cite{Olson:2011}.
The lexicode generation algorithm tests potential codewords and chooses the ones that satisfy an arbitrary constraint.
For the traditional version, this constraint is that if a codeword is to be selected, it should be at least a predetermined Hamming distance away from the previously chosen codewords~\cite{Conway:1986}.
This can be adapted to marker lexicode generation by also testing for the circular permutations of the codewords.

The described lexicode generation algorithm is an exhaustive search of an $n$-dimensional binary space, $n$ being the length of the codewords.
This corresponds to a time complexity of $O(2^n)$.
With $48$-bit long codewords, the algorithm requires an unreasonable amount of time to finish, even with a GPU implementation.
To overcome this problem, we took a two-step hierarchical approach.
We first generated 12-bit long codewords, then generated the full-length codewords as 4-combinations of these.

\begin{table}
	\centering
	\caption{STag library sizes with respective minimum Hamming distances.}
	\tabulinesep=1.2mm
	\begin{tabu}{l c c c c c c c}
		\textbf{Minimum HD} & 11 & 13 & 15 & 17 & 19 & 21 & 23 \\
		\hline
		\textbf{Library Size} & 22,309 & 2,884 & 766 & 157 & 38 & 12 & 6 \\ 
	\end{tabu}
	
	\label{tab:libsize}
\end{table}

We generated libraries with different sizes and Hamming distances to accommodate for a variety of applications.
Smaller libraries with higher minimum Hamming distances can be used for small-scale AR applications where occlusion is common.
Larger libraries are more suitable for navigation through large interior spaces or inventory applications.
The resulting library sizes are presented in Table~\ref{tab:libsize}.

The main performance metric for a coding scheme is the amount of error correction it can provide for various library sizes.
Bit error ratio (BER) is the number of error bits divided by the number of total bits.
A marker library that can correct $0.1$~BER will work when at most $10\%$ of the coding area is read incorrectly.
By using this metric, we can compare libraries with different code lengths.

\begin{figure}
	\centering
	\includegraphics[width=0.7\columnwidth]{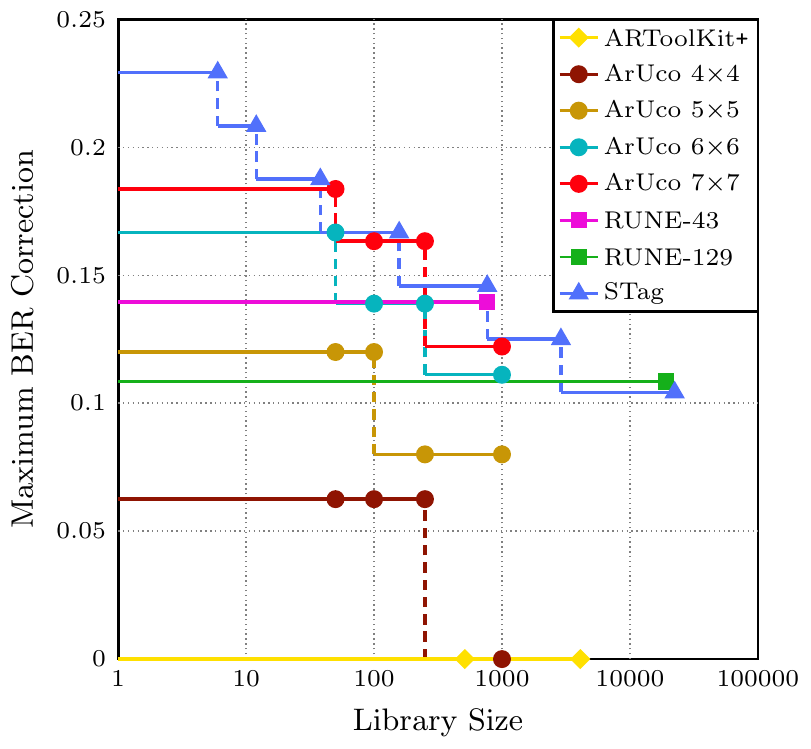}
	\caption{Marker libraries with different sizes and maximum bit error ratio~(BER) correction capabilities.}
	\label{fig:ber}
\end{figure}

See Figure~\ref{fig:ber} for a comparison of ARToolKit\texttt{+}~\cite{Wagner:2007}, ArUco~\cite{Garrido:2016}, RUNE-Tag~\cite{Bergamasco:2016}, and STag libraries.
Marker variants with different code lengths are plotted separately.
The horizontal axis is in logarithmic scale, because improving error correction capability reduces library size exponentially.

ARToolKit\texttt{+} implementation provides no error correction capability.
ArUco variants grow more efficient with longer codewords.
Both RUNE-Tag libraries are efficient, but they do not provide the best performance across the whole scale.
ArUco $7{\times}7$, RUNE-$129$ and STag libraries lie on the same line, which implies a theoretical upper bound.
STag libraries cover a large range of library sizes, and provide the best or near-best performance.

%-------------------------------------------------------------------------
\section{Detection Algorithm}

\begin{figure}
	\centering	
	\includegraphics[width=\textwidth]{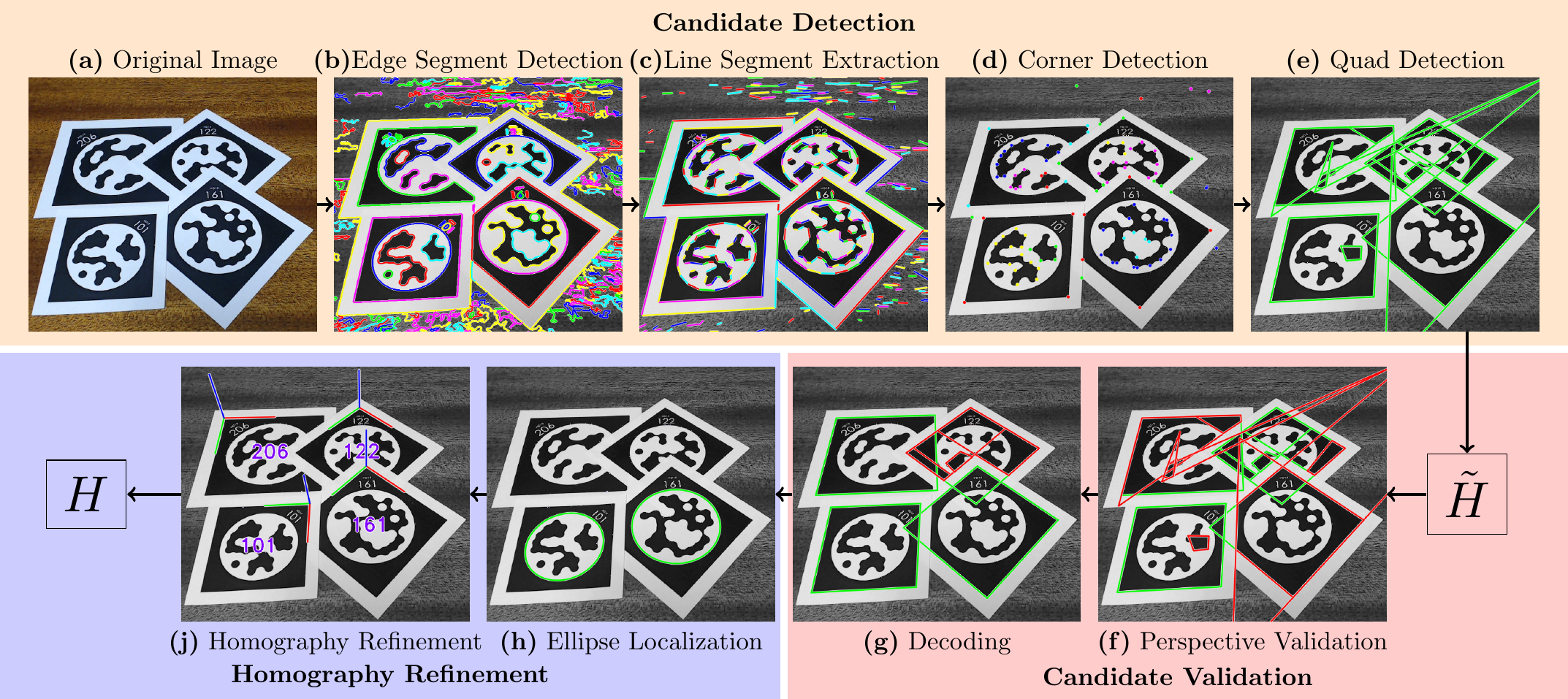}	
	\caption{Flowchart of the detection algorithm for the proposed marker system (best viewed in color).}
	\label{fig:Flowchart}
\end{figure}

The detection algorithm is composed of three main parts, as shown in Figure~\ref{fig:Flowchart}.
In the candidate detection part, which is discussed in Section~\ref{sec:CandidateDetection}, we detect all quads in the image.
These candidates are validated firstly by their shape, and secondly by their encoding, which is discussed in Section~\ref{sec:CandidateValidation}.
For the candidates that are validated to be markers, the computationally intensive homography refinement step is run, as described in Section~\ref{sec:PoseRefinement}.
After homography refinement, the pose (i.e., $[R|\textbf{t}]$) can be estimated using \cite{Schweighofer:2006}.

%-------------------------------------------------------------------------
\subsection{Candidate Detection}
\label{sec:CandidateDetection}

The first step of the candidate detection algorithm is to detect edge segments as contiguous arrays of pixels, which significantly eases further processing. 
For this task, we employ the EDPF algorithm~\cite{Akinlar:2012}, which determines anchor points in the image, and joins them by aiming to maximize the total gradient response along its path.
Then, it validates these edge chains according to the Helmholtz principle to obtain native and well-localized edge segments.
The edge segments detected in a sample image are shown in Figure~\hyperref[fig:Flowchart]{7b}.

As the next step, we process the edge segments to find the linear borders of the markers.
This is done by fitting a line to the beginning pixels of an edge segment, and extending this line as long as the following edge pixels are on it.
This operation is repeated until all edge segments are processed.
This approach avoids repeated line fitting operations, which become expensive with a large number of iterations~\cite{Akinlar:2011}.
In Figure~\hyperref[fig:Flowchart]{7c}, the extracted line segments are shown.

Corners are detected by intersecting the line segments extracted consecutively from the same edge segment.
Detected corners for the sample image is shown in Figure~\hyperref[fig:Flowchart]{7d}.
Following this, three consecutive corners are used to detect the quads on the image.
This allows the detection algorithm to detect quads even when a corner is occluded.
See all quads detected on the sample image in Figure~\hyperref[fig:Flowchart]{7e}.

\begin{figure}
	\centering
	\includegraphics[width=0.5\columnwidth]{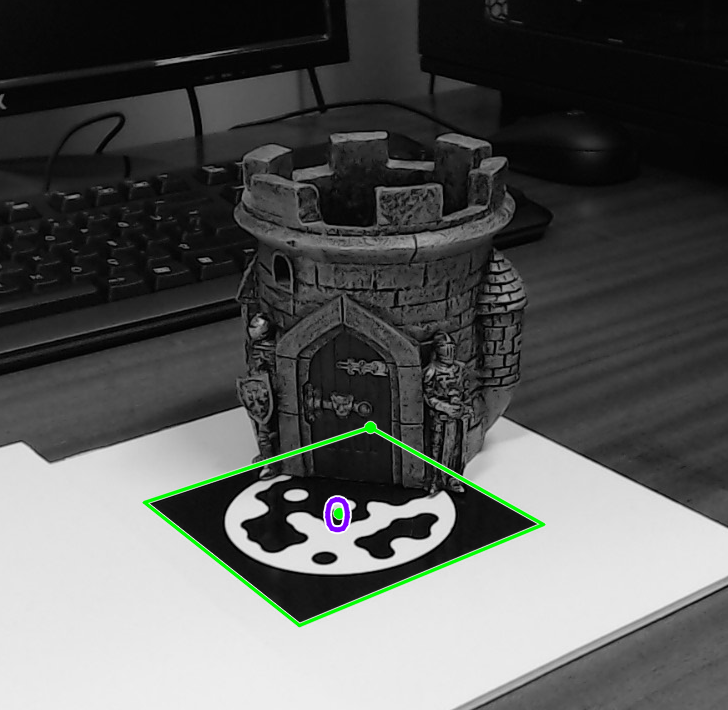}
	\caption{The marker is detected successfully under heavy occlusion.}
	\label{fig:occlusion}
\end{figure}

As seen in Figure~\ref{fig:ber}, STag libraries provide the highest BER correction capability.
This means that the markers can be decoded robustly against occlusion.
Together with the detection algorithm described in this section, we can detect markers even under significant occlusion.
See Figure~\ref{fig:occlusion} and the supplementary video for examples.

%-------------------------------------------------------------------------
\subsection{Candidate Validation}
\label{sec:CandidateValidation}

We propose a novel candidate validation algorithm that utilizes the candidate shape.
Specifically, we eliminate candidates based on the imposed perspective distortion.
Here, \textit{perspective distortion} is in reference to the aspect of a projective transformation excluding the part that can be represented as an affine transformation.
The objective is to eliminate as much false candidates as possible, which will decrease the false positive rate.

%Traditional marker detection algorithms validate candidates solely by decoding which waste CPU time.
%We propose a novel candidate validation method which eliminates false marker candidates without decoding them.
%Thus we both decrease false detection rates and execution time.
%In our method, we aim to infer whether the detected quad is a \textit{valid} projection of a square shape in real world.
%In other words, we try to find if the modal deviation of the candidate quad is due to the \textit{perspective distortion} or not.

Depth ($\alpha$) is the distance from a point to the camera.
The relative depth ($\alpha_{rel}$) of an object is the ratio of the largest depth ($\alpha_{max}$) to the smallest depth ($\alpha_{min}$) of the object.
The effect of perspective distortion becomes increasingly apparent as $\alpha_{rel}$ of the object increases.
If we can find $\alpha_{rel}$ of the candidates, we can eliminate them accordingly, assuming $\alpha_{rel}$ will be bounded by some constraints.
%Therefore, our approach can also be seen as thresholding the candidates based on the imposed perspective distortion.

\begin{figure}
	\centering
	\includegraphics[width=0.8\columnwidth]{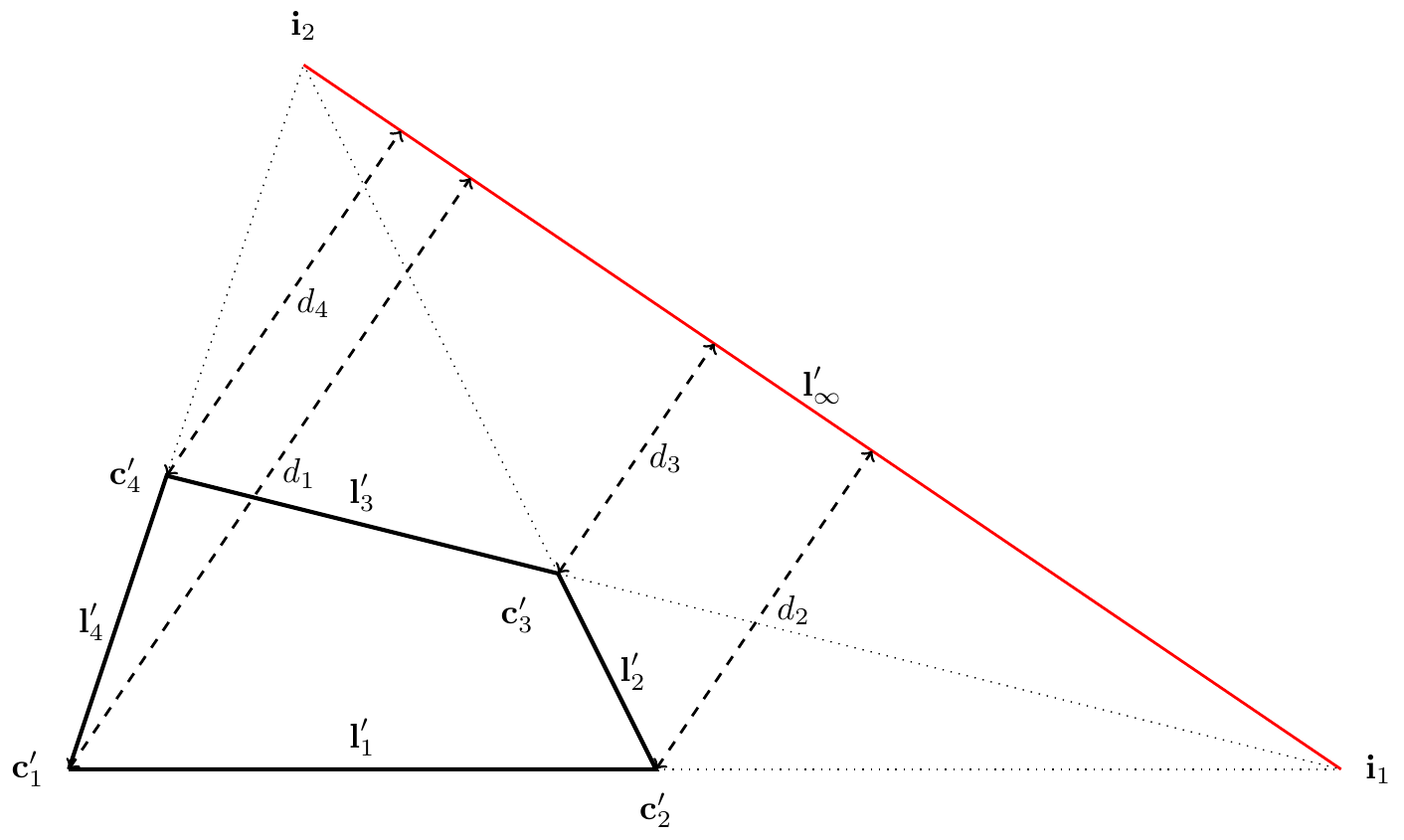}
	\caption{A marker candidate with corners $\textbf{c}_i'$ and vertices $\textbf{l}_i'$.
		The projective transformation have projected the line at infinity to $\textbf{l}_{\infty}'$, which can be found by using the intersections $\textbf{i}_1$ and $\textbf{i}_2$.
		Distances from $\textbf{c}_i'$ to $\textbf{l}_{\infty}'$ are shown with $d_i$.}
	\label{fig:perspective}
\end{figure}

Both opposite edges of a square are parallel, and lines extending along these parallel edges intersect at two distinct points at infinity.
The line at infinity ($\textbf{l}_{\infty}$) passes through these two points at infinity.
When a projective transformation is applied, each of these points at infinity are projected on a respective finite vanishing point.
The line passing through this vanishing point pair is the image of $\textbf{l}_{\infty}$ (see Figure~\ref{fig:perspective}).
The projection of $\textbf{l}_{\infty}$ can also be found by using the respective homography matrix~\cite{Hartley:2003}.
The points closest to and farthest to $\textbf{l}_{\infty}'$ will be two of the four corners of the quad.
The distance between $\textbf{l}_{\infty}'$ and points on the object ($d_i$) has a linearly negative relationship with the depth of the said point ($\alpha_i$)~\cite{Sparr:1992}.
Using this relationship, we can find the relative depth of a planar object using Equation~\ref{eq:RelDist} ($k$ is a scalar, refer to Figure~\ref{fig:perspective} for further notation).
\begin{equation}
\alpha_{rel} = \frac{\alpha_{max}}{\alpha_{min}} = \frac{kd_{min}^{-1}}{kd_{max}^{-1}} =\frac{d_{max}}{d_{min}}
\label{eq:RelDist}
\end{equation}

$\alpha_{rel}$ of the projected quad indicates the amount of perspective distortion it underwent.
A threshold is needed to eliminate candidates based on this metric.
%Let us assume that the markers in the application will not approach the camera more than 20~cm, which will give us $\alpha_{min}$.
%If the markers are sized $10$~cm$\times10$ cm, we can calculate $\alpha_{max}$ as $20+ 10\sqrt{2}$ cm.
Assume that $10$~cm$\times10$~cm markers will be used in an application and none of them will approach the camera more than 20~cm ($\alpha_{min}$).
We can calculate $\alpha_{max}$ to be $20+ 10\sqrt{2}$ cm, which results in the maximum $\alpha_{rel}$ to be $1.707$.
Therefore, we can safely eliminate all quads that have a larger $\alpha_{rel}$ than this threshold.
If the eliminated quads indicated with red in Figure~\hyperref[fig:Flowchart]{7f} are examined, it is seen that only the ones that can not be a valid projection of a square shape are eliminated.
We are going to show the benefit of this validation step experimentally in Section~\ref{sec:markerless}.

We estimate the homography matrix, $H$, of a candidate by using its corners as correspondence points~\cite{Hartley:2003}.
There are four ways of matching the corresponding corners, of which an arbitrary one is used.
The projections of bit representations are sampled using this homography to obtain the embedded codeword.

We keep all codewords in the library and their circular rotations in a list.
The read codeword is compared with each element in the list by XORing and doing a population count.
If the Hamming distance between the read codeword and a codeword in the list is smaller than or equal to the maximum number of bits to be corrected, the respective rotation and ID is associated with the candidate.
If the arbitrarily chosen rotation is found to be incorrect, the homography is updated with the correct correspondence.
If the codeword can not be found in the ID list, then that marker is also eliminated after decoding (see the red quads in Figure~\hyperref[fig:Flowchart]{7g}).

%-------------------------------------------------------------------------
\subsection{Homography Refinement}
\label{sec:PoseRefinement}

\begin{figure}
	\centering
	\subfloat[]
	{
		\label{fig:arcfind1}
		\includegraphics[width=0.24\columnwidth]{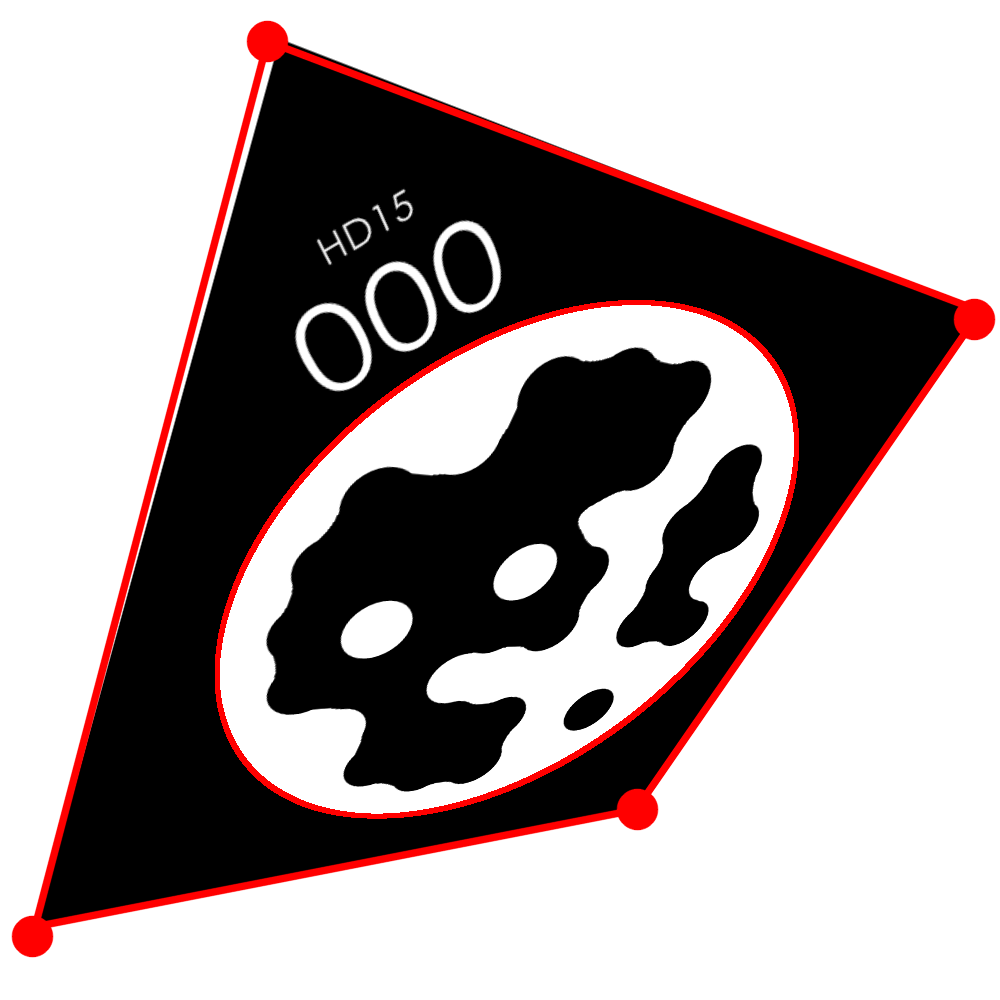}
	}
	\subfloat[]
	{
		\label{fig:arcfind2}
		\includegraphics[width=0.24\columnwidth]{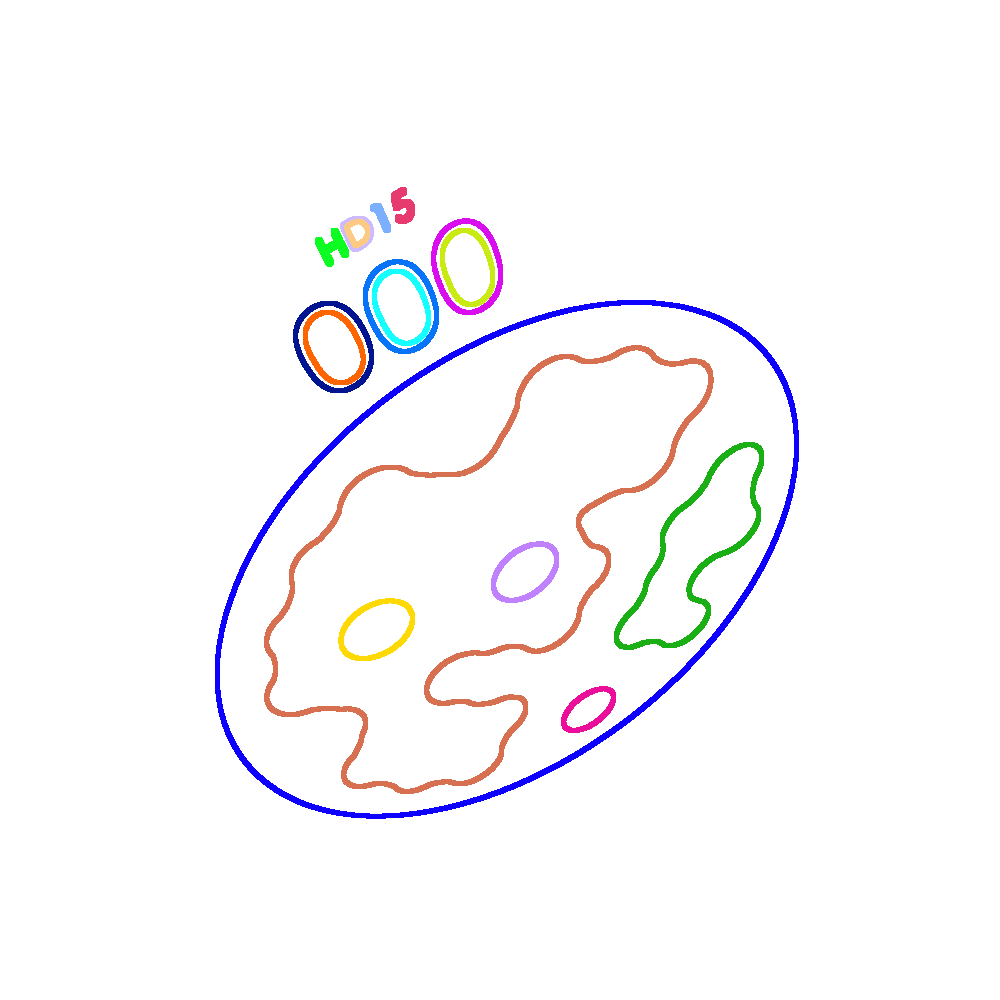}
	}
	\subfloat[]
	{
		\label{fig:arcfind3}
		\includegraphics[width=0.24\columnwidth]{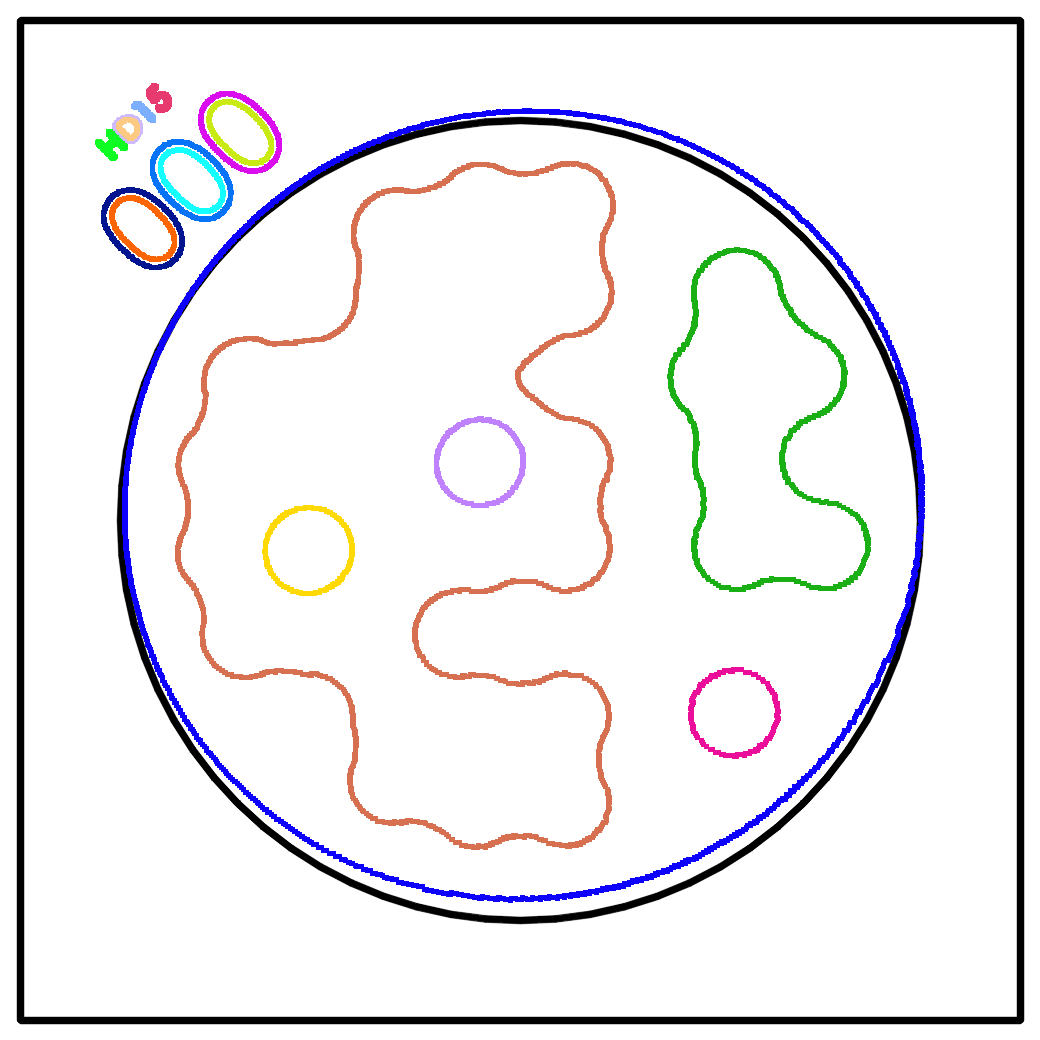}
	}
	\subfloat[]
	{
		\label{fig:arcfind4}
		\includegraphics[width=0.24\columnwidth]{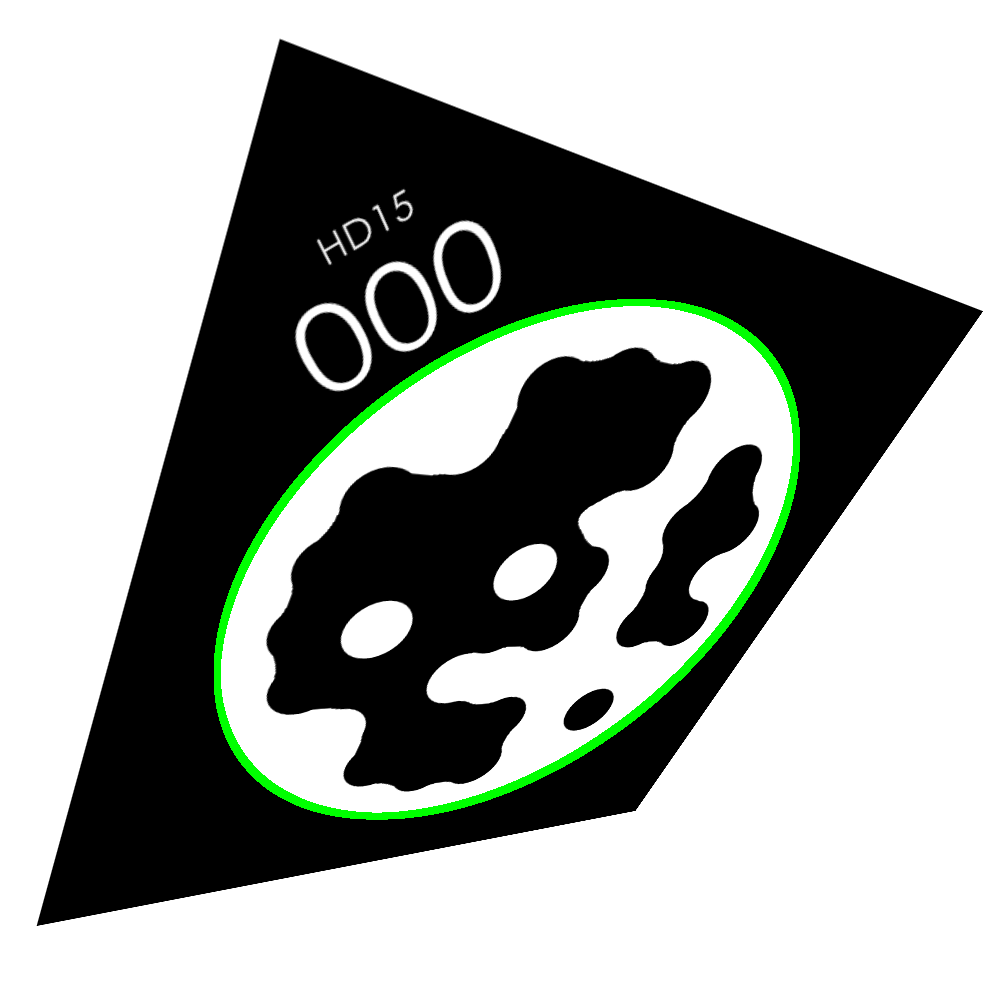}
	}
	\caption{(a)~A marker with incorrectly localized corners and the respective expected ellipse location.
		(b)~Edge segment loops inside the marker detection.
		(c)~Edge segment loops backprojected to the marker plane.
		(d)~The ellipse fitted to the chosen edge segment loop.
	}
	\label{fig:arcfind}
\end{figure}

At this point, we have a set of markers that are validated by decoding.
We are going to localize the inner circular border as an ellipse, and utilize the conic information to refine the estimated homography.
Let us illustrate the ellipse localization process.
In Figure~\ref{fig:arcfind1}, we present an example with exaggerated localization error for better viewing.
We first find the edge segment loops inside the marker detection (see Figure~\ref{fig:arcfind2}) and backproject them to the marker plane (see Figure~\ref{fig:arcfind3}).
It is expected that among the edge segment loops, the one from the border ellipse will be the most similar to the inner circular border.
We estimate this similarity by sparsely sampling the distances between the inner circular border and the back projected edge segment loops.
An ellipse is fitted to the edge segment loop whose back projection is the most similar to the inner circular border~\cite{Fitzgibbon:1999} (see Figure~\ref{fig:arcfind4}).

We detect the absence of a suitable edge segment loop by thresholding the similarity between the backprojected edge segment loops and the inner circular border.
Since this metric is calculated in the marker plane, the shape and size of the detection is normalized.
Thus, the threshold acts equivalently in all poses.
If a suitable ellipse could not be detected, homography refinement is omitted (e.g., markers \#122 and \#206 in Figure~\hyperref[fig:Flowchart]{7j}).
This is a deliberate design choice, because we have observed that homography refinement with a partially occluded ellipse does not benefit stability, and in some cases even degrades it.

%To localize the ellipse, we back project the edge segment loops inside the marker detection to the marker plane.
%We find the back projected edge segment loop that is the most similar to the inner circular border by sparsely sampling the distances.
%An ellipse is fitted to the edge segment loop whose back projection is the most similar to the inner circular border~\cite{Fitzgibbon:1999} (see Figure~\hyperref[fig:Flowchart]{7h}).
%We detect the absence of a suitable edge segment loop by thresholding the similarity between the back projected edge segment loops and the inner circular border.
%Since this metric is calculated in the marker plane, the shape and size of the detection is normalized.
%Thus, the threshold acts equivalently in all poses.

%In this section, we will be considering circles and ellipses as conic sections.
%Let us start with a brief overview of 2D projections of conic sections.
%See~\cite{Hartley:2003} for more details as needed.
In the following discussion, we are going to be considering circles and ellipses as conic sections, which are represented as matrices.
Capital letters refer to matrices, bold-face letters refer to vectors, and plain letters refer to scalars.
Where $\textbf{x}$ is a point in homogeneous coordinates and $\textbf{x}'$ is its projection, the 2D transformation is represented as such~\cite{Hartley:2003}:
\begin{equation}
\textbf{x}'=H\textbf{x}
\label{eq:pointH}
\end{equation}
Under the same transformation as Equation~\ref{eq:pointH}, the projection of a conic section is commonly represented as the following two equivalent equations:
\begin{equation}
C'=H^{-T}CH^{-1}
\label{eq:conicH1}
\end{equation}
\begin{equation}
C=H^{T}C'H
\label{eq:conicH2}
\end{equation}
Let $C$ be the inner circular border, whose projection is localized as an ellipse $C'$.
When we back project $C'$ to the marker plane with Equation~\ref{eq:conicH2}, we get an ellipse $\tilde{C}$, which does not coincide with $C$ perfectly.
This is because the initial homography estimated using the marker corners is slightly incorrect.
Our approach is going to be to optimize $H$ to make $C$ and $\tilde{C}$ as similar as possible. 

To optimize $H$, we need a similarity metric between the ellipse $\tilde{C}$ and the circle $C$.
Where $a$ is the semi-major axis, $b$ is the semi-minor axis , $(e_x, e_y)$ is the center of $\tilde{C}$, and $r$ is the radius, $(c_x, c_y)$ is the center of $C$, respectively, the metric we propose is:
\begin{equation}
\epsilon=\sqrt{(e_x - c_x)^2 + (e_y - c_y)^2 + (a -r)^2 + (b - r)^2}
\label{eq:diffmetric}
\end{equation}
These distances are in the marker plane, thus they are normalized for pose.

A single circle--ellipse correspondence is not adequate to estimate the homography of a marker.
We use the homography estimated using the corners as a starting point, and minimize $\epsilon$ in Equation~\ref{eq:diffmetric} using the Nelder--Mead method~\cite{Nelder:1965}.
This way, the previously estimated homography is refined such that the ellipse detection is back projected directly on the inner circular border.
Since the ellipse is localized more correctly than the marker corners, this improves the stability of the localization.
See Section~\ref{sec:poserefbenefit} for an experiment that demonstrates the benefits of homography refinement.

%-------------------------------------------------------------------------
\subsection{Running Time and Parallelization}
\label{sec:RunningTimeAndParallelization}

\begin{table}
	\centering
	\caption{Running times of detection algorithm steps in milliseconds.}\vspace{1.5cm}
	\tabulinesep=1.2mm
	\begin{tabu}{p{0.1\textwidth} p{0.1\textwidth} p{0.1\textwidth} p{0.1\textwidth} p{0.1\textwidth} p{0.1\textwidth} p{0.1\textwidth} p{0.1\textwidth}}
		\begin{rotate}{30} EDPF~\cite{Akinlar:2012} \end{rotate} & \begin{rotate}{30} EDLines~\cite{Akinlar:2011} \end{rotate} & \begin{rotate}{30} Candidate Detection \end{rotate} & \begin{rotate}{30} Candidate Validation \end{rotate} & \begin{rotate}{30} Ellipse Localization \end{rotate} & \begin{rotate}{30} Homography Refinement \end{rotate} & \begin{rotate}{30} \textbf{Total} \end{rotate} \\
		\hline
		22.6 & 1.0 & 0.7 & 0.6 & 1.1 & 0.8 & \textbf{26.8} \\ 
	\end{tabu}
	
	\label{tab:parallel}
\end{table}

Since marker detection algorithms have to run in real time, it is important for them to both run fast and be parallelizable in case a further speed up is needed.
In this section, we are going to analyze the proposed detection algorithm in this regard.
See Table~\ref{tab:parallel} for the steps of the detection algorithm for a $1280{\times}720$ image with a cluttered scene containing a single marker.

The first thing that meets the eye is that EDPF constitutes more than 80\% of the running time.
In fact, the total running time is higher than Table~\ref{tab:runningtime} simply because of the larger number of edge segments extracted from the background clutter.
As the name implies, the Edge Drawing algorithm is partly sequential.
However, a CUDA implementation of this algorithm has resulted in up to ${\times}12$ speedup even with older model GPUs~\cite{Ozsen:2012}.
This would correspond to a ${\times}4.4$ speedup in the STag detection algorithm by itself.

The rest of the detection algorithm is considerably more parallelizable than the edge segment detection step.
EDLines and candidate detection process edge segments independently.
Considering that there are hundreds of edge segments in a typical image, processing of these can be parallelized with arbitrary granularity.
Following this, candidate validation operates on candidates independently, and around 3.7 candidates can be expected on a typical scene (derived from the results in Table~\ref{tab:falsepositives}).
Finally, ellipse localization and homography refinement only operate on successfully decoded markers.
In the case that there are multiple markers on the scene, these steps are also easily parallelizable.
Alternatively, these two final steps can be omitted if there are many markers on the scene, as a large number of correspondences from multiple markers would already provide a stable pose.

%-------------------------------------------------------------------------
\section{Experiments}

We compared the proposed marker system with ARToolKit\texttt{+}~\cite{Wagner:2007}, ArUco~\cite{Garrido:2014} and RUNE-Tag~\cite{Bergamasco:2016} in a series of experiments.
These marker systems are comparable with ours, in that they both provide large libraries and allow pose estimation with a single marker.
The literature uses warped synthetic images for pose experiments, which is not an adequate approximation.
All our experiments are done on real images.

%-------------------------------------------------------------------------
\subsection{Detection}
\label{sec:markerlessdetection}

\begin{table}
	\centering
	\caption{Marker library characteristics and detection performance at the Indoor Scene Recognition Dataset~\cite{Quattoni:2009}.}
	
	\resizebox{\textwidth}{!}{
		\begin{tabu}{l c c c c c c c c}
			\textbf{Marker System} &
			\textbf{Marker Library} &
			\textbf{Bits} &
			\parbox{1.9cm}{\centering \textbf{Error Correction}} &
			\textbf{Library Size} &
			\parbox{2.5cm}{\centering \textbf{Number of Candidates}} &
			\parbox{2.5cm}{\centering \textbf{Number of False Positives}} &
			\parbox{2.5cm}{\centering \textbf{Validation Failure Probability${}^1$}}\\ \hline
			\multirow{2}{*}{ARToolKit\texttt{+}~\cite{Wagner:2007}} & simple-id & 36 & 0 & 512 & 11739 & 0 & N/A \\
			& BCH-id & 36 & 0 & 4096 & 11739 & 3 & 6.2E-8 \\ \hline
			\multirow{17}{*}{ArUco~\cite{Garrido:2014,Garrido:2016}} & ARUCO\_ORIGINAL & 16 & 0 & 1024 & 317741 & 785 & 2.4E-06 \\
			& 4X4\_50 & 16 & 1 & 50 & 317741 & 2349 & 7.4E-05 \\
			& 4X4\_100 & 16 & 1 & 100 & 317741 & 3250 & 5.1E-05 \\
			& 4X4\_250 & 16 & 1 & 250 & 317741 & 13018 & 8.2E-05 \\
			& 4X4\_1000 & 16 & 0 & 1000 & 317741 & 1875 & 5.9E-06 \\
			& 5X5\_50 & 25 & 3 & 50 & 317741 & 269 & 2.1E-06 \\
			& 5X5\_100 & 25 & 3 & 100 & 317741 & 934 & 3.7E-06 \\
			& 5X5\_250 & 25 & 2 & 250 & 317741 & 294 & 9.3E-07 \\
			& 5X5\_1000 & 25 & 2 & 1000 & 317741 & 1821 & 1.4E-06 \\
			& 6X6\_50 & 36 & 6 & 50 & 317741 & 979 & 9.6E-07 \\
			& 6X6\_100 & 36 & 5 & 100 & 317741 & 177 & 1.7E-07 \\
			& 6X6\_250 & 36 & 5 & 250 & 317741 & 939 & 3.7E-07 \\
			& 6X6\_1000 & 36 & 4 & 1000 & 317741 & 212 & 4.2E-08 \\
			& 7X7\_50 & 49 & 9 & 50 & 317741 & 27 & 3.3E-09 \\
			& 7X7\_100 & 49 & 8 & 100 & 317741 & 2 & 2.5E-10 \\
			& 7X7\_250 & 49 & 8 & 250 & 317741 & 112 & 5.53E-09 \\
			& 7X7\_1000 & 49 & 6 & 1000 & 317741 & 4 & 2.0E-10 \\ \hline
			\multirow{2}{*}{RUNE-Tag~\cite{Bergamasco:2011}} & Rune-43${}^2$ & 43 & 6 & 762 & N/A & N/A & N/A \\
			& Rune-129 & 129 & 14 & 17000 & 0 & 0 & N/A \\ \hline
			\multirow{7}{*}{STag} & HD11 & 48 & 5 & 22309 & 57893 & 7 & 1.7E-10 \\
			& HD13 & 48 & 6 & 2884 & 57893 & 6 & 5.6E-10 \\
			& HD15 & 48 & 7 & 766 & 57893 & 23 & 4.1E-09 \\
			& HD17 & 48 & 8 & 157 & 57893 & 11 & 4.7E-09 \\
			& HD19 & 48 & 9 & 38 & 57893 & 5 & 4.4E-09 \\
			& HD21 & 48 & 11 & 12 & 57893 & 2 & 2.8E-09 \\
			& HD23 & 48 & 13 & 6 & 57893 & 38 & 5.3E-08 \\ 
			\multicolumn{8}{l}{{${}^{1}$ \text{This metric shows the probability of a false positive for a marker library size of $1$ with no error correction. Its objective is to assess}}} \\
			\multicolumn{8}{l}{false positive rates without penalizing larger marker libraries and higher error correction capabilities. Lower is better.} \\
			\multicolumn{8}{l}{${}^{2}$ \text{Rune-43 is not provided by the authors}} 
		\end{tabu}
	}
	\label{tab:falsepositives}
\end{table}

We used an indoor scene recognition dataset with $15{,}620$ images from $67$ categories to represent markerless scenes~\cite{Quattoni:2009}.
See Table~\ref{tab:falsepositives} for library characteristics and respective numbers of candidates and false positives.
We used the error correction capability of each library to its fullest extent.
While both ArUco and STag are trying to detect square markers, STag detects significantly less false candidates.
RUNE-Tag detects no candidates in the entire dataset.

Looking at the number of false positives in Table~\ref{tab:falsepositives}, we see that STag returns few false positives even with extreme error correction.
Compared to STag, ArUco has returned more false positives.
To eliminate the effect of the number of candidates, amount of error correction and library size to the discussion, we developed a metric named Validation Failure Probability:
\begin{equation}
\resizebox{\columnwidth}{!}{
	$\text{Validation~Failure~Probability} = \frac{\text{No.~False~Positives}}{\text{No.~Candidates} \times \text{Library~Size} \times 2^{\frac{\text{Error~Correction}}{\text{Bits}}}}$
}
\end{equation}
By this metric, we see that the reduced number of false positives from STag is not only due to the lower number of candidates, but also a better performing marker library generation scheme.
Moreover, the validation performance is consistent for different library sizes and error correction rates.

\begin{figure}
	\centering
	\includegraphics[width=0.7\columnwidth]{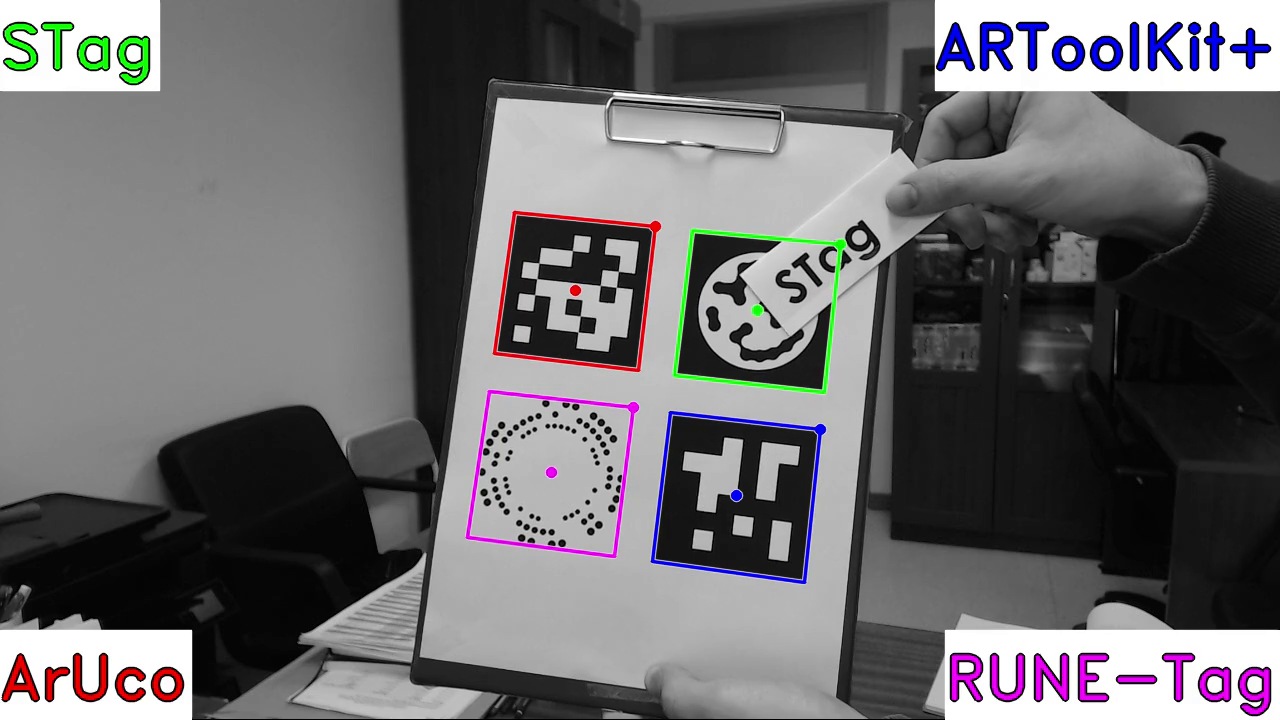}
	\caption{A frame from the supplementary video that demonstrates detection performance.
		See \url{https://github.com/bbenligiray/stag} for the video.}
	\label{fig:supp}
\end{figure}

For a qualitative experiment about detection with markers on the scene, see the supplementary video (see Fig.~\ref{fig:supp}).
A marker of ARToolKit\texttt{+}, ArUco, RUNE-Tag and STag are printed on a sheet.
They are moved around, rotated and occluded to compare the detection algorithms of the respective systems.
RUNE-Tag cannot be detected when the markers are further away.
The temporal tracking of ARToolKit\texttt{+} results in mislocalization when the marker is moving rapidly.
ArUco and ARToolKit\texttt{+} are not very robust against occlusion.
STag is detected consistently throughout the video, is robust against occlusion, and is localized stably due to homography refinement.
Note that the processing was done offline, as RUNE-Tag cannot be run with this frame rate in real time.

%-------------------------------------------------------------------------
\subsection{Perspective Validation}
\label{sec:markerless}

\begin{table}
	\centering
	\caption{The number of false positive detections with maximum error correction in the indoor scene recognition dataset~\cite{Quattoni:2009} with and without the candidate validation method described in Section~\ref{sec:CandidateValidation}.}
	\tabulinesep=1.2mm
	\begin{tabu}{l c c c c c c c}
		\textbf{Minimum HD} & 11 & 13 & 15 & 17 & 19 & 21 & 23 \\
		\hline
		\textbf{w/o validation} & 15 & 16 & 34 & 13 & 6 & 6 & 98 \\ 
		\textbf{w/ validation} & 7 & 6 & 23 & 11 & 5 & 2 & 38 \\ 
	\end{tabu}
	\label{tab:val}
\end{table}

Let us demonstrate the benefits of the proposed shape-based candidate validation method described in Section~\ref{sec:CandidateValidation} using the same indoor scene recognition dataset~\cite{Quattoni:2009}.
A total of $94{,}427$ quads were detected in the entire dataset, and this number was reduced to $57{,}893$ after being validated based on perspective.
To clearly see the effect of this reduction, we ran the decoding algorithm with maximum error correction capability (half of the minimum Hamming distance), which typically returns a large number of false positives.
See Table~\ref{tab:val} for the results.
Indeed, a significant portion of the eliminated candidates were going to be detected as false positives.

%-------------------------------------------------------------------------
\subsection{Setup for Stability Experiments}
\label{sec:setup}

For stability experiments, we captured images of a stationary marker using a stationary camera.
Ideally, the localization and pose differences between these images should be zero.
Although the marker detection algorithms are deterministic, imaging noise causes localization differences between the images, which result in pose differences.
This reflects to the end user as instability and jitter in pose.
We calculated the standard deviation of these differences to compare the stability of different marker systems.

\begin{figure}
	\centering
	\includegraphics[width=0.5\columnwidth]{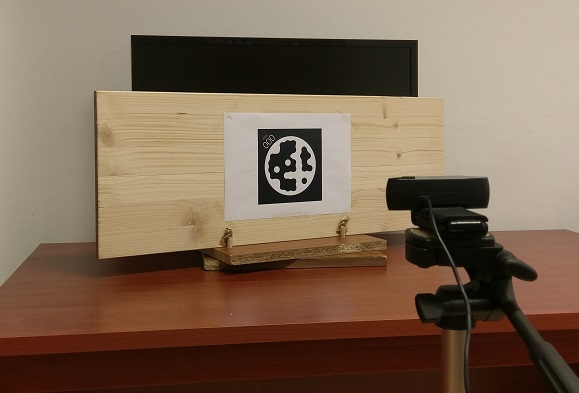}
	\caption{A jig is built to manipulate the viewing angle without moving the camera.
		The marker printed on paper is affixed tightly to the jig by taping its corners.
		The webcam is mounted on a tripod at marker height, aimed straight at the center of the marker.
		The viewing angle and distance is estimated using additional markers.}
	\label{fig:expsetup}
\end{figure}

See Figure~\ref{fig:expsetup} for the experimental setup of the stability experiments.
A $15$~cm-wide marker printed on paper is tightly affixed to a jig built for easy pose manipulation.
We chose to print the markers on paper rather than rigid material such as polystyrene tiles, because this better represents the typical usage scenario.
A Logitech C920 webcam is mounted on a tripod at marker level and aimed directly at the center of the marker.
The viewing angle and distance are adjusted by keeping the camera stationary and manually manipulating the jig, taking the pose estimated from additional markers as reference.
The pose is set up to $1^{\circ}$ and 1~cm accuracy.
For each marker system, distance and viewing angle, 1000 frames were captured (i.e., each point in the following figures is the standard deviation over 1000 frames).

The camera's focus and exposure were set manually, and kept the same for all experiments.
Anti-flicker was on and lens distortion was rectified beforehand.
The captured images were grayscale and of $1280{\times}720$ resolution.
The lighting was completely artificial, coming from overhead sources.
Since the camera did not move, its relative position with respect to the lighting sources did not change.
In addition, we made sure that there was no significant air flow in the environment during the experiments that could move the markers.

%-------------------------------------------------------------------------
\subsection{Benefits of Homography Refinement}
\label{sec:poserefbenefit}

\begin{figure}
	\centering
	\includegraphics[width=0.8\columnwidth]{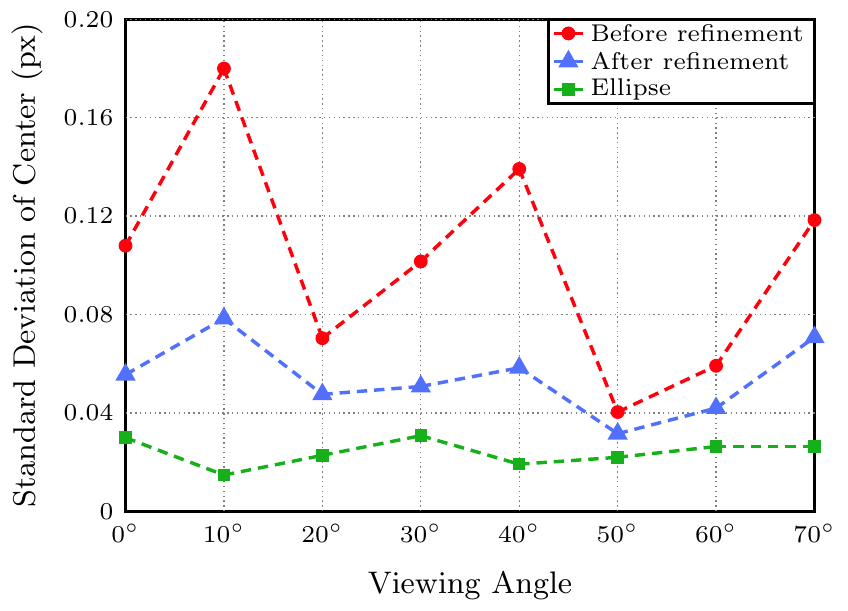}
	\caption{Localization stability before refinement and after refinement.
		Before refinement, marker center stability is very low.
		Refining with the well-localized ellipse results in a significant improvement in stability.}
	\label{fig:benefit}
\end{figure}

In this section, we are going to demonstrate that the circular border is localized more stably than the square border in real images, and the proposed homography refinement step improves localization stability.
For this specific experiment, a $10$~cm-wide marker is positioned $100$~cm away from the camera.
Images with a resolution of $1920 \times 1080$ are captured from a series of viewing angles.
The stability of the detection is represented by the standard deviation of the center localization.
The marker centers before and after refinement are localized by projecting the marker center on the image with the respective homography matrix.
See Figure~\ref{fig:benefit} for the results.
Before refinement, the marker center is unstable.
Moreover, the localization is not robust against the changes in viewing angles.
The center of the ellipse we have located is significantly more stable.
After homography refinement, the marker center gains stability across all viewing angles.

%-------------------------------------------------------------------------
\subsection{Pose Estimation Stability}
\label{sec:pes}

ArUco~\cite{Garrido:2014} and RUNE-Tag~\cite{Bergamasco:2016} uses OpenCV's Perspective-n-Point implementation to estimate the pose from a single marker.
For coplanar correspondence points, this implementation estimates the homography matrix, decomposes it as described in \cite{Zhang:2000} and refines it to minimize the projection error if there are redundantly many correspondence points.
The method described in \cite{Zhang:2000} requires the homography matrix to be estimated using at least two images where the orientation of the planar object is different.
With a single image of a planar object, the pose estimation has four different solutions of which two satisfy the cheirality constraints~\cite{Hartley:1998}.
Therefore, there is an ambiguity between two solutions.
Especially when the marker is seen from far away or an acute viewing angle, the estimated pose switches between the two possible solutions randomly, causing a significant amount of pose jitter.

ARToolkit\texttt{+} uses the method proposed in \cite{Schweighofer:2006} to find both possible solutions and select the one that returns the smaller reprojection error in the object-space.
We should note that in practice, this method does not always converge to the correct solution when only 4 marker corners are used, as we have observed jittering due to pose ambiguity while using it.
Nevertheless, it is obviously superior over traditional homography decomposition for difficult viewing conditions, which is why we have also used it to estimate the pose of a single marker for STag.

The estimated pose has 6 degrees of freedom, thus it is difficult to quantify the amount of jitter in the pose with a single metric.
For this reason, we investigated the jitter with different metrics.
Namely, we checked jitter in rotation, translation, and center localization.
To quantify the jitter in rotation, we found an average rotation across the frames.
Axis--angle representation of rotation is composed of a direction and a magnitude.
By using the magnitude part, we can represent the difference from the mean rotation, which was used to calculate the deviation in degrees.
Quantifying the jitter in translation is rather straightforward.
A mean translation vector is found for all frames, then the L2 distance from this mean is used to calculate the standard deviation in metric units.
Finally, the center of the marker is localized by using the camera matrix $K$ and the pose matrix $[R|\textbf{t}]$ to project the marker center on the image.
The standard deviation is found using the L2 distance on the image in pixel units.
Note that the pose ambiguity between the two possible solutions only causes jitter in rotation, and should not affect translation or marker center localization.

%-------------------------------------------------------------------------
\subsubsection{Different Viewing Angles}
\label{sec:DiffViewAngle}

\begin{table}
	\centering
	\caption{Number of false negative detections out of 1000 frames.}
	\resizebox{\textwidth}{!}{
		\begin{tabu}{l c c c c c c c c c c c c c c c c c c}
			\textbf{Viewing Angle} & $0^{\circ}$ & $5^{\circ}$ & $10^{\circ}$ & $15^{\circ}$ & $20^{\circ}$ & $25^{\circ}$ & $30^{\circ}$ & $35^{\circ}$ & $40^{\circ}$ & $45^{\circ}$ & $50^{\circ}$ & $55^{\circ}$ & $60^{\circ}$ & $65^{\circ}$ & $70^{\circ}$ & $75^{\circ}$ & $80^{\circ}$ & $85^{\circ}$ \\
			\hline
			ARToolKit\texttt{+} & 0 & 0 & 0 & 0 & 0 & 0 & 0 & 0 & 0 & 0 & 0 & 0 & 0 & 0 & \textcolor{red}{1000} & \textcolor{red}{1000} & \textcolor{red}{1000} & \textcolor{red}{1000} \\
			ArUco & 0 & 0 & 0 & 0 & 0 & 0 & 0 & 0 & 0 & 0 & 0 & 0 & 0 & 0 & 0 & 0 & 416 & \textcolor{red}{1000} \\
			RUNE-Tag & 0 & 0 & 0 & 0 & 420 & 0 & 11 & \textcolor{red}{970} & \textcolor{red}{1000} & \textcolor{red}{1000} & \textcolor{red}{1000} & \textcolor{red}{1000} & \textcolor{red}{1000} & \textcolor{red}{1000} & \textcolor{red}{1000} & \textcolor{red}{1000} & \textcolor{red}{1000} & \textcolor{red}{1000} \\
			STag & 0 & 0 & 0 & 0 & 0 & 0 & 0 & 0 & 0 & 0 & 0 & 0 & 0 & 0 & 0 & 0 & 4 & \textcolor{red}{1000} 
		\end{tabu}
	}
	\label{tab:falsenegative1}
\end{table}

For this experiment, we captured images of the markers from different viewing angles using the setup described in Section~\ref{sec:setup}.
Let us start by the detection accuracies (see Table~\ref{tab:falsenegative1}).
We highlighted the cases where more than half of the frames returned false negative detections in red, and we will not report stability metrics for these cases.
For the rest, we did not penalize the false negative detections in our stability calculations.

Both ArUco and STag are highly robust against difficult viewing angle conditions, yet STag performs significantly better from $80^{\circ}$.
ARToolKit\texttt{+} cannot detect markers seen from an angle of $70^{\circ}$ or more, yet is rather robust for lower viewing angles.
RUNE-Tag quickly stops being detectable as viewing angle increases, and is not very robust even in smaller viewing angles.
Seeing that RUNE-Tag is not prone to false positive detection (see Table~\ref{tab:falsepositives}), but fails to detect markers under suboptimal conditions, we can conclude that RUNE-Tag has high precision, but poor recall.

\begin{figure}
	\centering
	\subfloat[Rotation]
	{
		\label{fig:stabrot}
		\includegraphics[width=0.8\columnwidth]{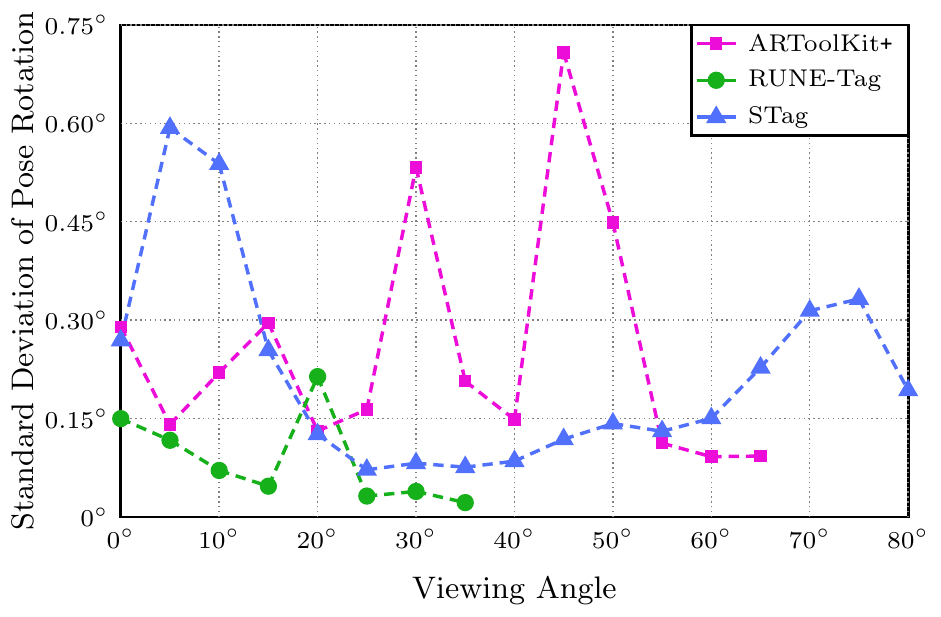}	
	}
	
	\subfloat[Translation]
	{
		\label{fig:stabtrans}
		\includegraphics[width=0.8\columnwidth]{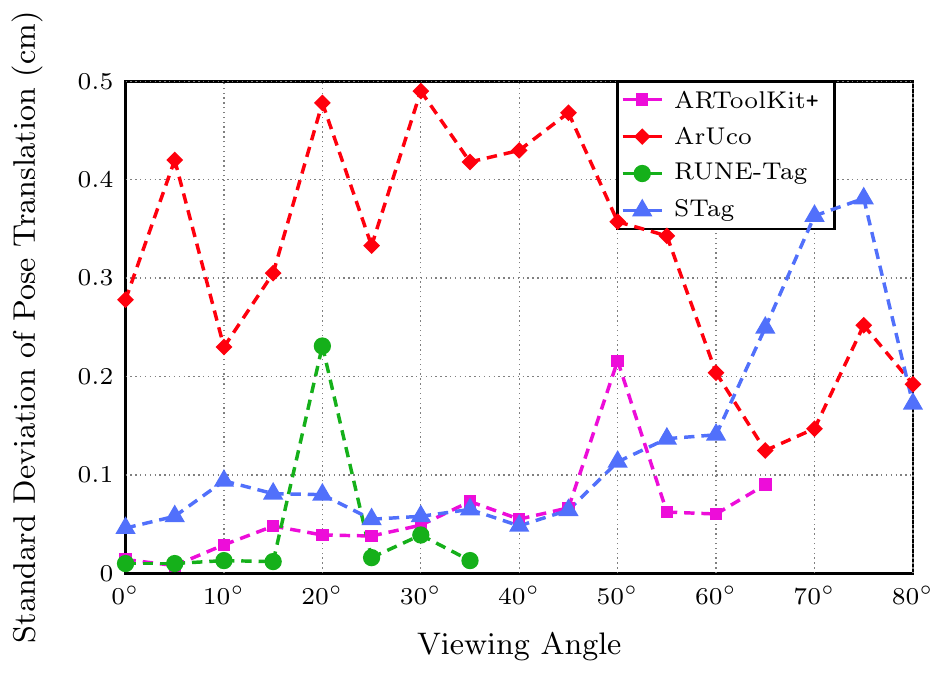}
	}
	\caption{Pose rotation and translation stability with different viewing angles.
		Standard deviation of pose rotation for ArUco was $11.70^{\circ}$ on average, which could not be shown in this scale.}
	\label{fig:stab1}
\end{figure}

See Figure~\ref{fig:stab1} for the rotation and translation stability across viewing angles.
The proposed marker system is both stable and detectable across viewing angles.
However, its stability suffers under more acute viewing angles both due to localization difficulties and pose ambiguity.
While RUNE-Tag achieves high stability due to many correspondence points, these fine dots also degrade easily, resulting in unreliable detection.
ARToolkit\texttt{+} localizes the marker corners coarsely, which causes its stability to depend on where the corner localizations fall on the quantization grid.
ArUco has considerable jitter both due to unstable localization (seen on Figure~\ref{fig:stabtrans}) and high pose ambiguity.
In fact, standard deviation of pose rotation for ArUco was very high ($11.70^{\circ}$ on average), which is why we were not able to illustrate it in Figure~\ref{fig:stabrot}.
This is to be expected, as the rotation difference between the two ambiguous poses tends to be a lot more than the jitter caused by localization instability.

\begin{figure}
	\centering
	\includegraphics[width=0.8\columnwidth]{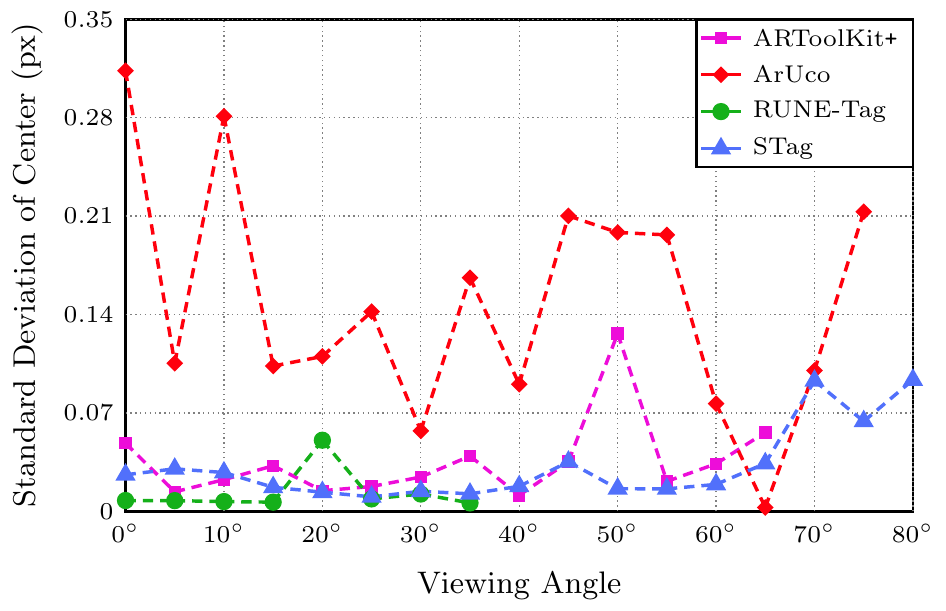}
	\caption{Localization stability with different viewing angles.}
	\label{fig:stabcenter}
\end{figure}

See Figure~\ref{fig:stabcenter} for the localization jitter given in pixel units.
We can see that these results are somewhat similar to the translation jitter results, as both are not affected by the pose ambiguity.
Again, the proposed system can be detected robustly, and its center can be localized stably.

\begin{figure*}
	\centering
	\subfloat[ARToolKit\texttt{+}]{\includegraphics[width=0.24\textwidth]{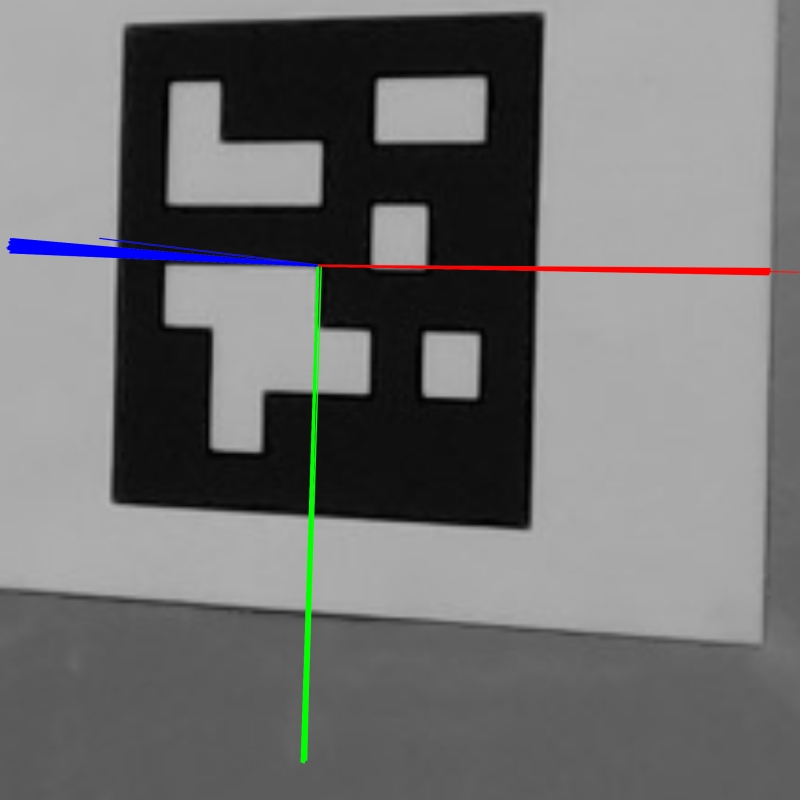}}
	~
	\subfloat[ArUco]{\includegraphics[width=0.24\textwidth]{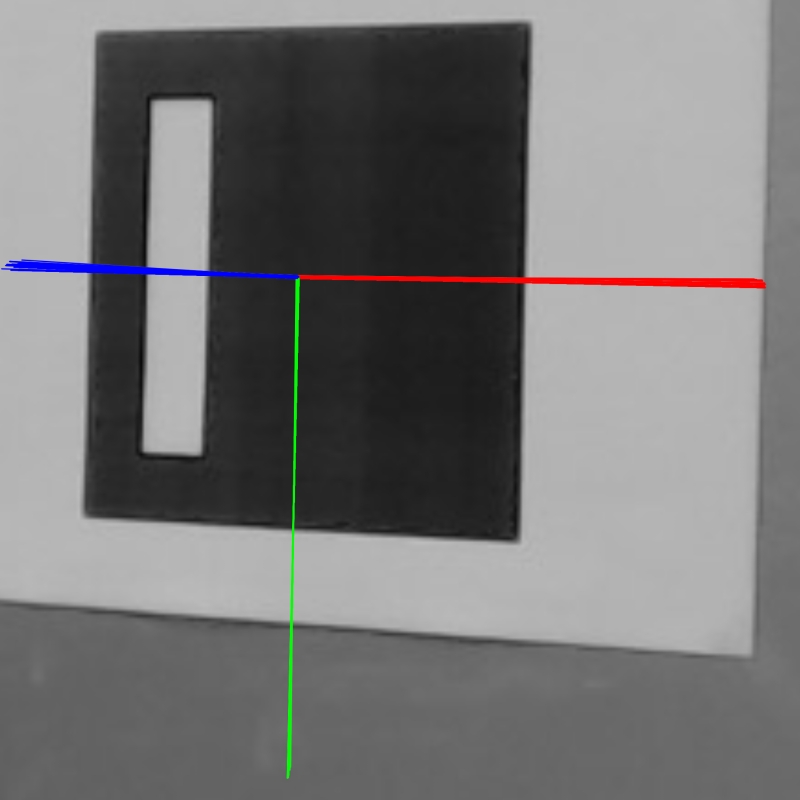}}
	~
	\subfloat[RUNE-Tag]{\includegraphics[width=0.24\textwidth]{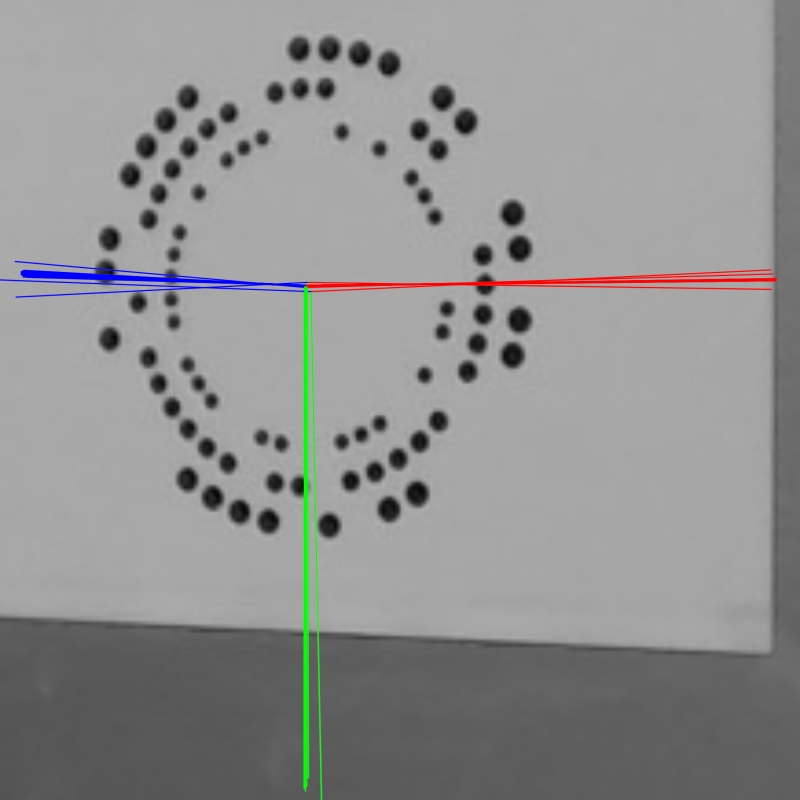}}
	~
	\subfloat[STag]{\includegraphics[width=0.24\textwidth]{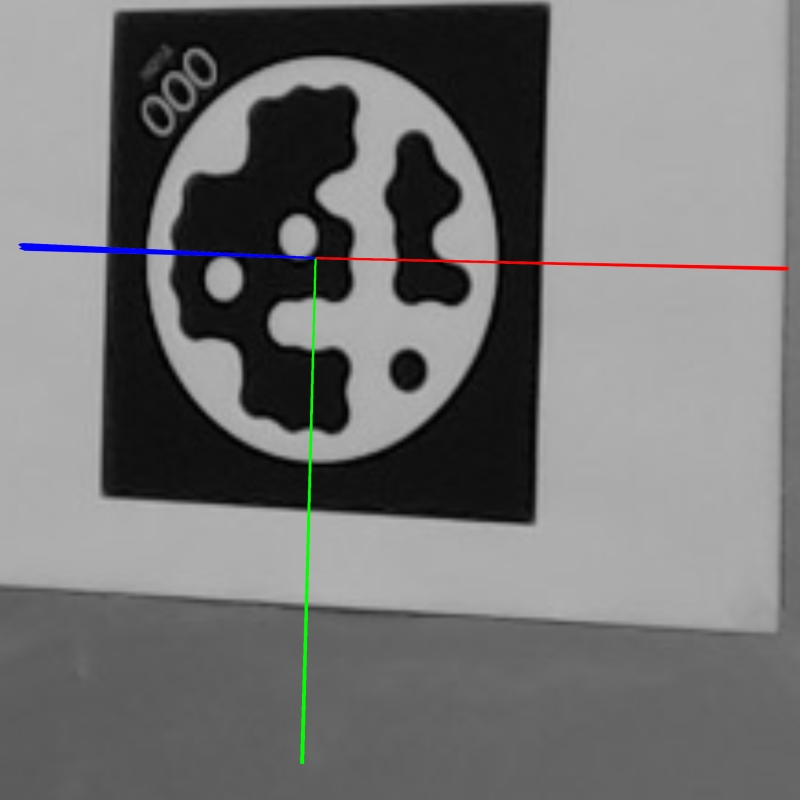}}
	\medskip
	\caption{$500$ pose axes are plotted on markers seen from $30^{\circ}$.
		The instability in the estimated pose causes the axes to appear scattered.
		The incorrect pose estimations due to the ambiguity have been eliminated manually.}
	\label{fig:axes}
\end{figure*}

Finally, let us illustrate the difference in stability qualitatively.
See Figure~\ref{fig:axes} for $500$ pose axes drawn on markers seen from $30^{\circ}$.
Note that we have eliminated incorrect poses due to pose ambiguity, so this figure only illustrates the jitter caused by localization instability.
When the estimated pose is stable, the axes will overlap, resulting in a single solid axis.
In contrast, an unstable pose will result in scattered axes.
The axes from ARToolKit\texttt{+} and ArUco seem to be scattered fairly uniformly.
The axes from RUNE-Tag seem quite solid, yet there are a few outliers.
On the other hand, the axes drawn on the proposed system appear to be overlapping, meaning that the proposed pose is stable.

%-------------------------------------------------------------------------
\subsubsection{Different Viewing Distances}
\label{sec:DiffViewDistance}

\begin{table}
	\centering
	\caption{Number of false negative detections out of 1000 frames.}
	\resizebox{\textwidth}{!}{
		\begin{tabu}{l c c c c c c c c c}
			\textbf{Viewing Distance (cm)} & 100 & 125 & 150 & 175 & 200 & 225 & 250 & 275 & 300 \\
			\hline
			ARToolKit\texttt{+} & 0 & 0 & 0 & 0 & 0 & 0 & 0 & 0 & 0 \\
			ArUco & 0 & 0 & 0 & 0 & 0 & 0 & 0 & 0 & 0 \\
			RUNE-Tag & 0 & \textcolor{red}{1000}  & \textcolor{red}{1000} & \textcolor{red}{1000} & \textcolor{red}{1000} & \textcolor{red}{1000} & \textcolor{red}{1000} & \textcolor{red}{1000} & \textcolor{red}{1000} \\
			STag & 0 & 0 & 0 & 0 & 0 & 0 & 0 & 0 & 0 \\
		\end{tabu}
	}
	\label{tab:falsenegative2}
\end{table}

We have repeated the experiment in Section~\ref{sec:DiffViewAngle} with a constant $0^{\circ}$ viewing angle and different viewing distances.
See Table~\ref{tab:falsenegative2} for detection performances.
Similar to the case with large viewing angles, RUNE-Tag cannot be detected when viewed from afar, as its coding circles appear too small to be detected.
Since the other systems are use a thick outer border for detection, they can be detected from a distance.

\begin{figure}
	\centering
	\subfloat[Rotation]
	{
		\label{fig:stabrot2}
		\includegraphics[width=0.8\columnwidth]{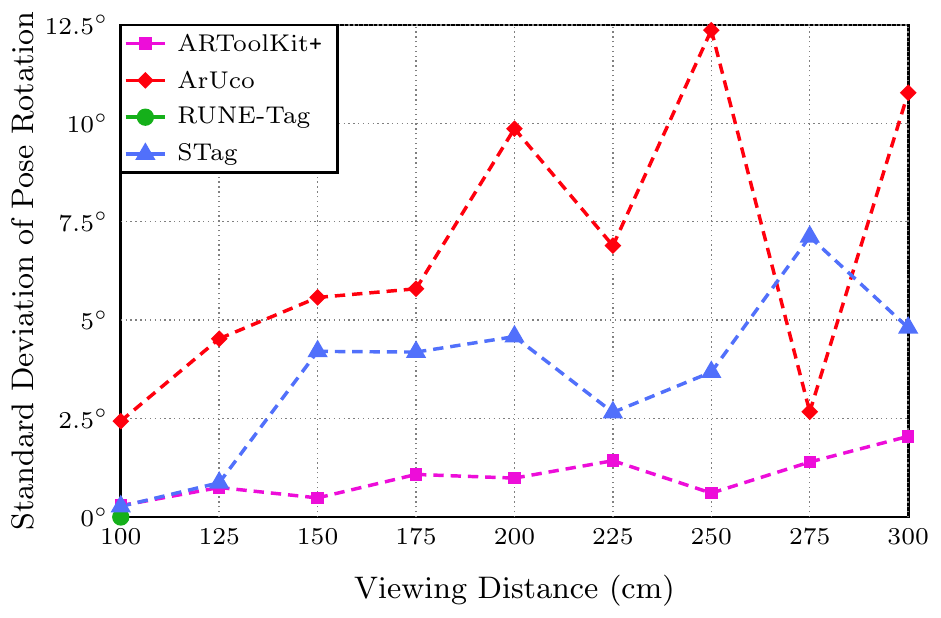}	
	}
	
	\subfloat[Translation]
	{
		\label{fig:stabtrans2}
		\includegraphics[width=0.8\columnwidth]{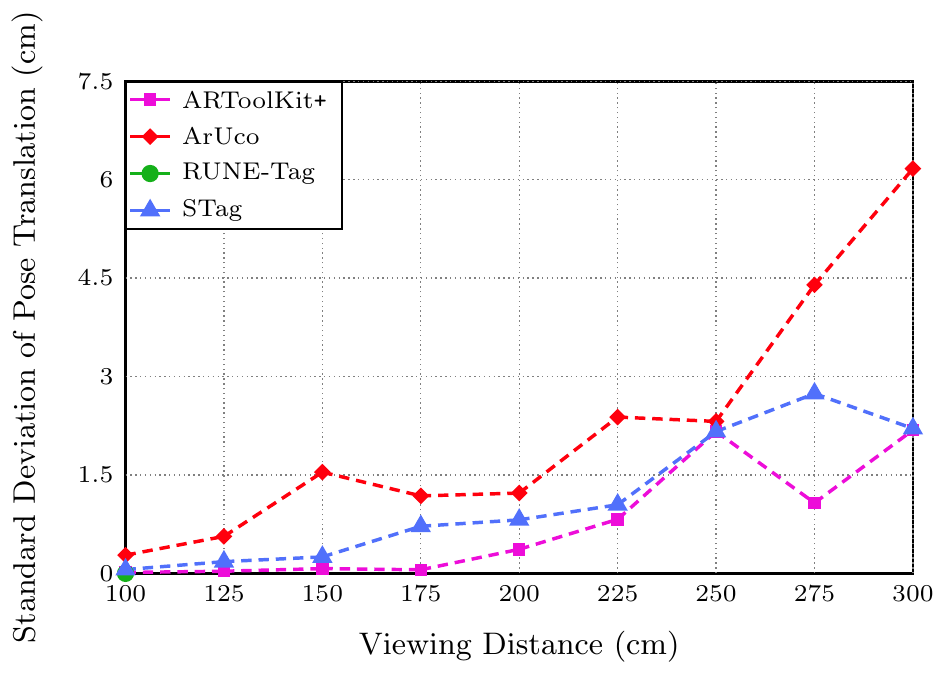}
	}
	\caption{Pose rotation and translation stability with different viewing distances.}
	\label{fig:stab2}
\end{figure}

See Figure~\ref{fig:stab2} for pose stability across viewing distances.
Unlike changing viewing distance, pose ambiguity becomes the dominant factor in pose stability, which is apparent from the rather large standard deviations in Figure~\ref{fig:stabrot2}.
ARToolkit\texttt{+} and STag are affected less because they use \cite{Schweighofer:2006} to estimate the pose.
ARToolkit\texttt{+} is affected even less than STag, as its corners are localized in a more coarse manner, which seems to be an advantage in the case where the marker appears smaller.
Compared to pose rotation standard deviations, pose translation standard deviations are much more comparable for all methods (see Figure~\ref{fig:stabtrans2}).

\begin{figure}
	\centering
	\includegraphics[width=0.8\columnwidth]{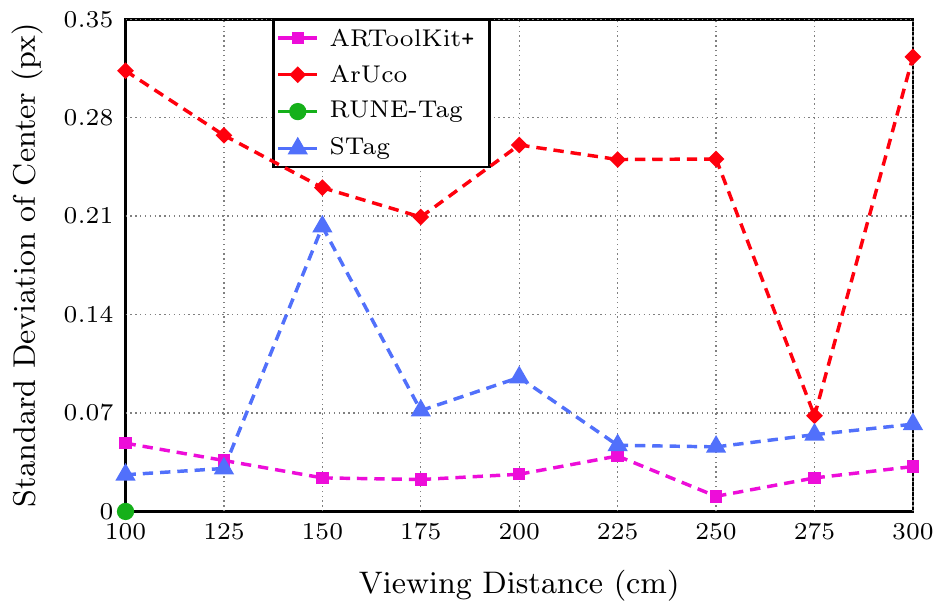}
	\caption{Localization stability with different viewing distances.}
	\label{fig:stabcenter2}
\end{figure}

Finally, we localized the marker centers using $[R|\textbf{t}]$.
It is interesting to see that the marker center stability metrics in Figure~\ref{fig:stabcenter} and Figure~\ref{fig:stabcenter2} are more similar compared to stabilities in pose.
This means that using the markers from a large distance does not degrade their localization stability as much as it degrades their pose stability.

%-------------------------------------------------------------------------
\subsection{Running Times}

\begin{table}
	\centering
	\caption{Running times of each algorithm with $1280 \times 720$ images with a single marker on the scene.}
	\tabulinesep=1.2mm
	\begin{tabu}{l c c c c }
		& ARToolKit\texttt{+} & ArUco & RUNE-Tag & STag \\
		\hline
		\textbf{Running time (ms)} & 2.6 & 10.0 & 579.9 & 18.1
	\end{tabu}
	\label{tab:runningtime}
\end{table}

Marker detection algorithms tend to be embarrassingly parallelizable.
To nullify the effects of different parallel implementations, we used a single core of a $3.70$ GHz Intel Xeon processor to run the experiments.
See Table~\ref{tab:runningtime} for average running times per image for the experiment in Section~\ref{sec:pes}.
We can see that ARToolKit\texttt{+} is very fast, both ArUco and STag perform in real-time, while RUNE-Tag is considerably slower.

%-------------------------------------------------------------------------
\section{Conclusion}

The main contribution of the proposed marker system is improving stability with a homography refinement step.
This is achieved by estimating the homography with the outer square border and refining it with the inner circular border.
To our knowledge, this problem and the respective solution is unique in the literature in the way that a single conic correspondence is used, without additional projective constraints such as a point~\cite{Lopez:2002}.

The proposed solution is significantly more stable than ArUco~\cite{Garrido:2014}.
Since it does not depend on a large number of fine details, it is significantly more robust across viewing conditions and runs an order of magnitude faster than RUNE-Tag~\cite{Bergamasco:2016}.
However, we have seen that its localization characteristics are somewhat more prone to pose ambiguity compared to ARToolkit\texttt{+}~\cite{Wagner:2007}.
Considering this pose ambiguity is an important factor in pose estimation stability, future work should focus on marker designs that takes this issue into account.

For the multi-marker case, the inner circular borders can be used to estimate the pose with a multiple conic correspondence approach~\cite{Kannala:2006}.
The alternative we have proposed is using the ellipse centers as stable, yet inaccurate correspondences of marker centers.
This may cause some inaccuracy, but the stability of the point correspondences will yield an even more stable pose estimation.

%\clearpage
{\small
	\bibliographystyle{ieeetr}
	\bibliography{References}
}

\end{document}